\documentclass{article}

\usepackage{microtype}
\usepackage{graphicx}
\usepackage{subfigure}
\usepackage{booktabs} 

\usepackage{hyperref}

\usepackage[accepted]{icml2024}

\usepackage{amsmath}
\usepackage{amssymb}
\usepackage{mathtools}
\usepackage{amsthm}

\usepackage[capitalize,noabbrev]{cleveref}

\theoremstyle{plain}

\theoremstyle{definition}

\theoremstyle{remark}

\usepackage[textsize=tiny]{todonotes}

\usepackage{tikz}
\usetikzlibrary{arrows.meta}
\newcommand{\mypoint}[2]{\tikz[remember picture]{\node[inner sep=0pt](#1){$#2$};}}

\definecolor{minecolortau}{rgb}{0.858, 0.188, 0.478}

\usepackage{graphicx}		
\usepackage{amsmath,amssymb, bm}		
\usepackage{wrapfig}		
\usepackage{algorithm}		
\usepackage{algorithmic}	
\usepackage{subfigure}		
\usepackage{color}			
\usepackage{multicol}		
\usepackage{multirow} 		

\newcommand{\lbr}{\left[ }
\newcommand{\rbr}{\right] }

\newcommand{\bz}{\mathbf{z}}
\newcommand{\bv}{\mathbf{v}}
\newcommand{\btau}{{\bm{\tau}}}
\newcommand{\bepsilon}{{\bm{\epsilon}}}

\newcommand{\bs}{\mathbf{s}}
\newcommand{\ba}{\mathbf{a}}

\newcommand{\eg}{\emph{e.g.},\ }
\newcommand{\ie}{\emph{i.e.},\ }

\icmltitlerunning{DIffusion-guided DIversity for offline behavioral generation}

\begin{document}

\twocolumn[

\icmltitle{DIDI: Diffusion-Guided Diversity for Offline Behavioral Generation}



\icmlsetsymbol{equal}{*}

\begin{icmlauthorlist}
\icmlauthor{Jinxin Liu}{equal,westlake,zju}
\icmlauthor{Xinghong Guo}{equal,westlake}
\icmlauthor{Zifeng Zhuang}{westlake}
\icmlauthor{Donglin Wang}{westlake,wias}
\end{icmlauthorlist}

\icmlaffiliation{westlake}{School of Engineering, Westlake University, Hangzhou, China}
\icmlaffiliation{zju}{Zhejiang University, Hangzhou, China}
\icmlaffiliation{wias}{Institute of Advanced Technology, Westlake Institute for Advanced Study, Hangzhou, China}

\icmlcorrespondingauthor{Jinxin Liu}{liujinxin@westlake.edu.cn}
\icmlcorrespondingauthor{Donglin Wang}{wangdonglin@westlake.edu.cn}

\icmlkeywords{Diffusion, Reinforcement Learning, Diversity, Skill Learning}

\vskip 0.3in
]



\printAffiliationsAndNotice{\icmlEqualContribution} 

\begin{abstract}

In this paper, we propose a novel approach called \textbf{DI}ffusion-guided \textbf{DI}versity (\textbf{DIDI}) for offline behavioral generation. The goal of DIDI is to learn a diverse set of skills from a mixture of label-free offline data. We achieve this by leveraging diffusion probabilistic models as priors to guide the learning process and regularize the policy. By optimizing a joint objective that incorporates diversity and diffusion-guided regularization, we encourage the emergence of diverse behaviors while maintaining the similarity to the offline data. Experimental results in four decision-making domains (Push, Kitchen, Humanoid, and D4RL tasks) show that DIDI is effective in discovering diverse and discriminative skills. We also introduce skill stitching and skill interpolation, which highlight the generalist nature of the learned skill space. Further, by incorporating an extrinsic reward function, DIDI enables reward-guided behavior generation, facilitating the learning of diverse and optimal behaviors from sub-optimal data.


\end{abstract}

\section{Introduction}

Offline reinforcement learning (RL) has shown great promise in enabling agents to learn from past experiences without further interaction with the environment~\citep{levine2020offline, brandfonbrener2021offline, janner2021offline}. Naturally, it eliminates the need for time-consuming and costly online exploration and enables learning from pre-collected large datasets. However, one inherent requirement of this formulation is that the offline data must be labeled with rewards, which guide the learning process of the policy. In practice, a significant portion of datasets is often collected without reward labels, posing challenges in learning a useful policy from such a reward-free dataset, particularly when the offline data is suboptimal or noisy.

Moreover, offline real-world data is typically collected from a mixture of different data-collecting policies, resulting in datasets that exhibit multimodality and diversity~\citep{wang2022diffusion, shafiullah2022behavior, chen2022latent}. Learning directly from such datasets may lead to suboptimal performance or bias the learned policy towards a specific behavior. Recent studies have explored incorporating additional contextual variables~\citep{zhou2021plas, liu2023design} or utilizing powerful model architectures~\citep{chi2023diffusion, furuta2021generalized} to learn a set of behaviors from the dataset. However, a major limitation of these methods is the lack of encouragement for the emergence of diverse behaviors. In practical applications, the ability to command the diversity of behaviors is often desired, rather than solely generating a set of optional behaviors.

To address these challenges, we propose a novel approach called {DIDI} ({DI}ffusion-guided {DI}versity) for offline behavioral generation. The objective of DIDI is to learn optimal \textit{and} diverse behaviors from a mixture of label-free offline data. To achieve controllable behavioral generation, we introduce a contextual policy that can be commanded to produce a specific behavior. The learning of the contextual policy is guided by a diffusion probabilistic model acting as a regularization prior.

Yet, supervising the contextual policy with diffusion model is not trivial, since we cannot directly obtain $<$contextual input,  behavior target$>$ pairs from the diffusion model. To this end, we draw inspiration from online unsupervised RL and employ a three-step process: First, we command the contextual policy to produce a pseudo-behavior. Then, we use the diffusion model to \textit{relabel} the pseudo-behavior and obtain a proxy-target. Finally, using the relabeled $<$\textit{contextual input, proxy-target}$>$ 
, we can supervise the learning of the contextual policy. 
The key insight is that the relabeling (noising and denoising) process of the diffusion model allows us to obtain a relabeled proxy-target, effectively bringing it back to the offline data distribution and thus eliminating the potential out-of-distribution (OOD) issues in offline settings.

We empirically evaluate our DIDI approach in four decision-making domains: Push, Kitchen, Humanoid, and D4RL tasks. Across various action/observation spaces, our results demonstrate that DIDI successfully discovers diverse and discriminative skills. Compared to alternative baselines, DIDI generates more diverse behaviors and achieves superior performance. Additionally, we showcase the generalist nature of the learned skill space by illustrating skill stitching and interpolation. Finally, assuming the availability of extrinsic rewards, we show that DIDI can produce diverse and optimal behaviors. 

The contribution of this paper can be summarized as follows: 
\textbf{1)} We propose DIDI, a novel approach for offline behavioral generation that utilizes a diffusion probabilistic model as a prior to guiding the learning of a contextual policy. 
\textbf{2)} Through extensive experiments, we demonstrate the effectiveness of DIDI in discovering diverse and discriminative skills, as well as generating optimal behaviors with new extrinsic rewards. 
\textbf{3)} Furthermore, we illustrate the generalist nature of the learned skill space through skill stitching and interpolation.

\section{Related Work}

This work resides at the intersection of offline reinforcement learning (RL), unsupervised skill learning, and diffusion probabilistic models. In this section, we provide a concise overview of the related work in these domains. 

\subsection{Offline Reinforcement Learning}

{{Offline reinforcement learning (RL)}} refers to the setting where the agent learns from a fixed reward-labeled dataset without further interaction with the environment. To address the potential out-of-distribution (OOD) problem, recent works have introduced various designs to ensure the learned policy aligns with offline data distribution. These designs range from policy constraints~\citep{fujimoto2021minimalist, wu2019behavior, wu2022supported} to value regularization~\citep{kumar2020conservative, kostrikov2021offline, liu2023beyond}, and from iterative optimization~\citep{kumar2019stabilizing, zhuang2023behavior, yu2021combo} to non-iterative frameworks~\citep{emmons2021rvs, chen2021decision, liu2023design, lai2023chipformer, zhuang2024reinformer}. However, these methods are limited to learning a single policy and rely on extrinsic reward labels. Alternatively, some works propose using expert demonstrations~\citep{zolna2020offline, liu2023ceil, kim2021demodice, liu2023clue, sun2023offline} or human preferences~\citep{shin2021offline, kang2023beyond} to guide offline policy learning. However, both of them require additional extrinsic supervision (using rewards, demonstrations, or preferences), which may not be available in practice. Moreover, the learned policy in these methods tends to be biased towards a single behavior, especially when the dataset contains multiple behaviors. In contrast, our method can learn a diverse set of behaviors from the dataset without requiring any extrinsic supervision.

\subsection{Unsupervised Skill Learning}

{{Unsupervised skill learning}} aims to learn a set of skills from unlabeled interactions. Typically, unsupervised skill learning methods can be categorized into two groups: online setting and offline formulation. In online RL, skill learning is often formulated through the lens of empowerment~\citep{laskin2022cic, liu2022learn, tian2021unsupervised, achiam2018variational, eysenbach2018diversity, sharma2019dynamics, tian2021independent, liu2021unsupervised}, which seeks to maximize the mutual information between the skill and the induced behavior. In contrast, offline skill learning is often formulated as a behavioral cloning problem~\citep{furuta2021generalized, chi2023diffusion}, which directly fits a policy from the dataset. However, such a formulation inherently imposes burdens to the fitting model, especially when the offline data is multi-modal or noisy. Also in the offline setting, our DIDI method explicitly inherits the empowerment objective and uses a diffusion probabilistic model to guide the empowerment optimization. Compared with existing offline learning methods that heavily rely on the fitting model itself, one key difference is that we explicitly introduce a behavioral diversity objective, actively encouraging the emergence of~diversity.

\subsection{Diffusion Probabilistic Models}

{{Diffusion probabilistic models}} have emerged as powerful tools for modeling complex data distributions, surpassing previous methods in terms of sample quality and diversity~\citep{tevet2022human, zhang2022motiondiffuse, zhu2023diffusion}. For decision-making tasks, recent works have shown that diffusion models can be used to model the policy/value network~\citep{wang2022diffusion, hansen2023idql}, serve as a planner~\citep{janner2022planning, liang2023adaptdiffuser, mishra2023generative}, or act as a data synthesizer~\citep{lu2023synthetic, yu2023scaling, chen2023genaug, he2023diffusion}. In our work, we employ a diffusion probabilistic model as a prior to guiding the learning of contextual policy, akin to (but extending beyond) the role of a data synthesizer.

\section{Preliminaries}

\subsection{Offline Reinforcement Learning}

Throughout this work, we consider reinforcement learning (RL) in the framework of Markov decision process (MDP) specified by the tuple $\mathcal{M} = (\mathcal{S}, \mathcal{A}, \mathcal{T}, r, p_0, \gamma)$, where $\mathcal{S}$ and $\mathcal{A}$ represent the state and action space respectively, $\mathcal{T}(\bs_{t+1}|\bs_t,\ba_t)$ denotes the transition dynamics, $r(\bs_t,\ba_t)$ denotes the reward function, $p_0(\bs_0)$ denotes the initial state distribution, and $\gamma$ denotes the discount factor. 
In an environment (MDP), 
the goal of RL is to learn a (stationary) policy $\pi_\theta(\ba_t|\bs_t)$ that maximizes the expected discounted return $\mathcal{J}(\pi_\theta) = \mathbb{E}_{\pi_\theta(\btau)}\left[ R(\btau) \right]$, where 
$R(\btau) := \sum_{t=0}^{T} \gamma^t r(\bs_t, \ba_t)$ denotes the discounted return of a trajectory $\btau := (\bs_0, \ba_0, \dots, \bs_T, \ba_T)$, and we overload notation $\pi_\theta(\btau)$ to represent the probability of trajectory $\btau$ generated by running policy $\pi_\theta(\ba_t|\bs_t)$ in the environment with transition dynamics $\mathcal{T}$: $\pi_\theta(\btau) = \pi_\theta \otimes \mathcal{T} :=p_0(\bs_0)\pi_\theta(\ba_0|\bs_0)\prod_{t=0}^{T-1}\mathcal{T}(\bs_{t+1}|\bs_t, \ba_t)\pi_\theta(\ba_{t+1}|\bs_{t+1})$.

In offline RL~\citep{levine2020offline}, the agent only has access to a static dataset $\mathcal{D}:=\{\btau  | \btau \sim \pi_\mathcal{D}(\btau)\}$ generated by one or more behavioral policies $\pi_\mathcal{D}$, and cannot interact further with the environment.  
Thus, model-based offline RL algorithms propose that we can learn a proxy transition model $\hat{\mathcal{T}}(\bs_{t+1}|\bs_t,\ba_t)=\arg\max_{\hat{\mathcal{T}}}\mathbb{E}_\mathcal{D}\left[ \log \hat{\mathcal{T}}(\bs_{t+1}|\bs_t,\ba_t) \right]$ 
and optimize the regularized objective:
\begin{align}\label{eq:model-based-offline-rl}
\max_{\pi_\theta,\hat{\mathcal{T}}} \mathbb{E}_{\pi_{\theta,\hat{\mathcal{T}}}(\btau)}\lbr R(\btau) + \log \pi_\mathcal{D}(\btau) - \log \pi_{\theta,\hat{\mathcal{T}}}(\btau) \rbr,
\end{align} 
where $\pi_{\theta,\hat{\mathcal{T}}}(\btau)= \pi_\theta \otimes \hat{\mathcal{T}}$ denotes the trajectory distribution generated by running policy $\pi_\theta(\ba|\bs)$ in the proxy environment $\hat{\mathcal{T}}(\bs_{t+1}|\bs_t, \ba_t)$. Briefly speaking, the additional regularization ($\log \pi_\mathcal{D}(\btau) - \log \pi_{\theta,\hat{\mathcal{T}}}(\btau)$) in Equation~\ref{eq:model-based-offline-rl} represents a KL divergence, encouraging the new rollout trajectory $\pi_{\theta,\hat{\mathcal{T}}}(\btau)$ does not deviate heavily from the offline data distribution $\pi_\mathcal{D}(\btau)$, and such divergence can be replaced by other measure instances. 

\subsection{Emergence of Online Diverse Behaviors}\label{sec:diverse-skills}

In this work, we consider learning a latent-conditioned policy $\pi_\theta(\ba_t|\bs_t,\bz)$, where the latent variable $\bz \in \mathbb{R}^d$ is drawn from a prior (skill) distribution $\bz \sim p(\bz)$. Then we define the joint latent-variable trajectory distribution (in online environment) $\pi_\theta(\btau, \bz) = p(\bz) \pi_\theta(\btau|\bz)$, where $\pi_\theta(\btau|\bz)=p_0(\bs_0)\pi_\theta(\ba_0|\bs_0,\bz)\prod_{t=0}^{T-1}\mathcal{T}(\bs_{t+1}|\bs_t, \ba_t)\pi_\theta(\ba_{t+1}|\bs_{t+1},\bz)$. 
To encourage the emergence of diverse behaviors, we maximize the mutual information between trajectories and latent variables~\citep{eysenbach2018diversity, sharma2019dynamics} along with the return~\citep{kumar2020one}:
\begin{align}
\max_{\pi_\theta}&\   I(\btau; \bz) + \mathbb{E}_{p(\bz),\pi_\theta(\btau|\bz)}\left[ R(\btau) \right]  \nonumber\\
&= \mathbb{E}_{p(\bz),\pi_\theta(\btau|\bz)} \left[ \log p(\bz|\btau) - \log p(\bz) + R(\btau) \right], \nonumber
\end{align}
If we remove the above $R(\btau)$ term, it corresponds to the pure unsupervised RL objective. 
To optimize above objective, we can approximate $p(\bz|\btau)$ with a learned discriminator network $q_\phi(\bz|\btau)$ and derive the evidence lower bound: 
\begin{align}\label{eq:elbo-divserse-skills}
\max_{\pi_\theta, q_\phi} \mathbb{E}_{p(\bz),\pi_\theta(\btau|\bz)} \left[ \log q_\phi(\bz|\btau) - \log p(\bz)  + R(\btau) \right]. 
\end{align}
Besides the trajectory-level diversity as described above, we can alternatively choose to maximize the mutual information between next-states and skills: $\max I(\bs_{t+1}; \bz|\bs_t)$, encouraging different skills, $\pi(\ba_t|\bs_t,\cdot)$, to visit different states, or to maximize $I(\bs_T; \bz)$, encouraging different skills to reach different final states. 

\subsection{Diffusion Probabilistic Model}

Diffusion models are a class of likelihood-based models that generate samples by gradually removing noise from a signal. Specifically, denoising diffusion probabilistic models (DDPMs,~\citet{ho2020denoising}) generate samples $\btau  := \btau^0$ by reversing a Gaussian noising process\footnote{Note that in this paper, we use \textit{superscript} $n \in [0, N]$ to denote the number of (forward or reverse) diffusion iterations, and use \textit{subscript} $t \in [0, T]$ to denote the time-step along a trajectory.} (\textit{superscript} $n \in [0, N]$ \textit{denotes the number of diffusion iteration}):
\begin{align}
p(\btau^N) &= \mathcal{N}(\textbf{0}, \textbf{I}), \nonumber \\
p_\theta(\btau^{n-1}|\btau^n) &= \mathcal{N}(\btau^{n-1}| {\mu}_\theta(\btau^n,n), \Sigma_\theta(\btau^n, n)), \label{eq:ddpm-reverse} \\
p_\theta(\btau^0) &= \int p(\tau^N)\prod_{n=1}^{N}p_\theta(\btau^{n-1}|\btau^n) \text{d}\btau^{1:N}, \nonumber
\end{align}
where ${\mu}_\theta(\btau^n,n)$ and $\Sigma_\theta(\btau^n, n)$ are learned neural networks. 
In general, we learn $\theta$ by maximizing a variational lower bound of the log likelihood $\mathbb{E}_{\pi_\mathcal{D}(\btau^0)}\left[ \log p_\theta(\btau^0) \right]$:
\begin{align}\label{eq:diffusion-elbo}
\mathbb{E}_{q(\btau^0,\btau^{1:N})}\left[ \log p_\theta(\btau^0, \btau^{1:N}) - \log q(\btau^{1:N}|\btau^0) \right],
\end{align}
where the expectation $q(\btau^0,\btau^{1:N}) := \pi_\mathcal{D}(\btau^0)q(\btau^{1:N}|\btau^0)$ is a forward Gaussian noising process. According to a predefined schedule $\beta_1,\dots,\beta_N$, DDPM sets the forward Gaussian noising process with $q(\btau^{1:N}|\btau^0) = \prod_{n=1}^{N}\mathcal{N}(\btau^{n}| \sqrt{1-\beta_n}\btau^{n-1}, \beta_n\textbf{I})$. 
With $\alpha_n := 1 - \beta_n$ and $\bar{\alpha}_n := \prod_{i=1}^{n}\alpha_i$, we can write the marginal distribution $q(\btau^n|\btau^0) = \mathcal{N}(\btau^n| \sqrt{\bar{\alpha}_n}\btau^0, (1-\bar{\alpha}_n)\textbf{I})$ and express $\btau^n$ as: $\btau^n(\btau^0, {\bepsilon}) = \sqrt{\bar{\alpha}_n}\btau^0 + (1-\bar{\alpha}_n)\bepsilon$, $\bepsilon\sim \mathcal{N}(\textbf{0}, \textbf{I})$. 
In implementation, instead of directly predict $\btau^0$ (maximizing Equation~\ref{eq:diffusion-elbo}), we can alternatively predict the noise $\bepsilon$ added to $\btau^0$, and  
learn the diffusion parameters $\theta$ with
\begin{align}\label{eq:diffusion-predict-noise}
\min_{\bepsilon_\theta} \mathbb{E}_{n, \btau^n, \bepsilon} \left[ \Vert \bepsilon - \bepsilon_\theta(\btau^n, n) \Vert^2 \right],
\end{align} 
where the expectation $\mathbb{E}_{n, \btau^n, \bepsilon}$ is specified by Equation~\ref{eq:diffusion-elbo} \citep{ho2020denoising}. 
At testing/inference, we can then use Equation~\ref{eq:ddpm-reverse} to conduct the generative process by progressively denoising a noisy input starting from $\btau^N \sim \mathcal{N}(\textbf{0}, \textbf{I})$.

\section{Diffusion-Guided Behavioral Diversity}

As described in Section \ref{sec:diverse-skills}, encouraging diversity (for the contextual policy $\pi_\theta(\ba_t|\bs_t,\bz)$) requires us to keep track of the mutual information between trajectories (or states) and latent variables $\bz$. In online RL, we can model the conditional distribution $\pi_\theta(\btau|\bz)$ with the actual rollout in the environment, \eg $\pi_\theta(\btau|\bz) = \pi_\theta \otimes \mathcal{T}$. 
However, we consider this problem in the context of an offline RL setting that prohibits the online interaction with the environment. 

Following the model-based offline RL formulation, one straightforward solution is to learn an additional dynamics model $\hat{\mathcal{T}}$ that substitutes the rollout distribution with $\pi_{\theta,\hat{\mathcal{T}}}(\btau|\bz) = \pi_\theta \otimes \hat{\mathcal{T}}$. 
In implementation, such solution (learning $\pi_{\theta}(\ba_t|\bs_t,\bz)$ with  Equation~\ref{eq:elbo-divserse-skills} over the proxy $\hat{\mathcal{T}}$) can easily exploit the inaccuracies of the learned $\hat{\mathcal{T}}$ and encounter unreasonable behaviors. Thus, typical model-based offline RL algorithms incorporate additional uncertainty estimation or conservatism into the policy training~\citep{yu2020mopo, kidambi2020morel}, while such uncertainty estimation or conservatism conflicts with our diversity objective. 

In this work, we advocate a {single} network for both policy learning and dynamics modeling, avoiding compounding rollout errors over an additional proxy dynamics model. 
We formulate such network as a sequence model, $\pi_\theta(\btau_t|\bs_t, \bz)$, where $\btau_t := [\bs_t, \ba_t, \bs_{t+1}, \dots, \bs_T, \ba_T]$. Here, the subscript $t$ of trajectory $\btau_t$ indicates the starting time of the trajectory. 
At testing, we then execute the first action $\ba_t$ of the predicted $\btau_t \sim \pi_\theta(\btau_t|\bs_t, \bz)$ at state $\bs_t$ given skill $\bz$. For simplicity of notation, we also denote such action selection by $\ba_t \sim \pi_\theta(\btau_t|\bs_t, \bz)$. 

\subsection{Planning with Diffusion Models}
\label{sec:planning_with_diffusion_models}

Following Diffuser~\citep{janner2022planning}, one can first learn a diffusion model  $\pi_\psi(\btau_t|\bs_t)$ to approximate the offline data distribution $\pi_\mathcal{D}(\btau_t)$ with Equation~\ref{eq:diffusion-predict-noise}. 
Then, at inference, we can model RL as conditional sampling over the learned diffusion model. 
Specifically, to encourage diverse behaviors, we approximate the reverse process (Equation~\ref{eq:ddpm-reverse}) with 
\begin{align}\label{eq:diffuser-denoising}
\pi_\theta(\btau_t|\bs_t,\bz) &:= p_\psi(\btau_t^0|\bs_t,\bz), \nonumber \\
p_\psi(\btau^{n-1}_t|\btau^n_t, \bs_t, \bz) &= \mathcal{N}\left( \btau^{n-1}_t| {\mu}_\psi+g\Sigma_\psi,\Sigma_\psi \right),
\end{align}
where $\mu_\theta$ and $\Sigma_\theta$ are the corresponding parameters in Equation~\ref{eq:ddpm-reverse}, and $g := \nabla_{\btau^n_t} \left( q_\phi(\bz|\btau^n_t) + R(\btau^n_t) \right)$ denotes the diversity and reward-maximizing guidance. 
Intuitively, adding the gradient $g$ to each reverse step will encourage the final $\btau^0_t$ incorporating both the diversity ($q_\phi$) and reward ($R$) information, and towards the desired behavior.  

\begin{figure}[t]
	\begin{center}
		\includegraphics[width=1.\columnwidth]{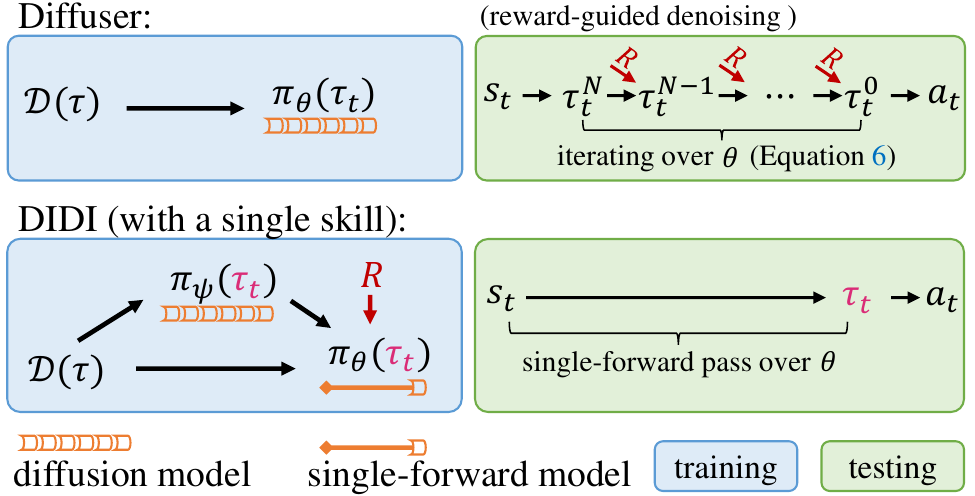}
	\end{center}
	\vspace{-7pt}
	\caption{Comparison between Diffuser~\citep{janner2022planning} and our DIDI (with a single skill, \ie assuming $p(\bz) = \delta(\bz)$).}
	\label{fig:diffuser-vs-didi} 
\end{figure}

However, iterating the reverse denoising process specified by Equation~\ref{eq:diffuser-denoising} leaves two main questions unanswered: 
\textbf{1)} It remains unclear how to obtain a pre-trained skill discriminator $q_\phi$ to guide the denoising process. 
Note that training $q_\phi$ is inherently different from the training of $R(\btau^n)$ that is orthogonal to the diffusion training. 
To derive an optimal discriminator $q_\phi(\bz|\btau_t)$, we need a meaningful trajectory $\btau_t$, which in turn depends on the performance of $q_\phi(\bz|\btau_t)$ that will guide the conditional diffusion sampling (Equation~\ref{eq:diffuser-denoising}). 
\textbf{2)} Performing action inference requires performing iterative denoising process (Figure~\ref{fig:diffuser-vs-didi} \textit{top}), which hinders their use in some real-time tasks. Even though \citet{janner2022planning} propose the warm-starting diffusion for faster planning, compared to the standard single forward-pass though inference networks, which still requires a large number of forward and backward (calculating gradient $g$) computations in all. To relax the burden of inference time, setting too small denoising step will suffer a clear performance~degradation.  

In the next subsection, we will discuss how we can address the above two challenges by incorporating pre-trained diffusion models (as priors) into the diversity-guided and reward-maximizing objective, and deriving a well-shaped policy network $\pi_\theta(\btau_t|\bs_t, \bz)$ that can directly produce decision action through a single forward-pass at inference, as shown in Figure~\ref{fig:diffuser-vs-didi} \textit{bottom}. 

\subsection{Diffusion Probabilistic Models as Priors}

Recall that our goal is learning diverse skills/behaviors, formulated by contextual policy $ \pi_\theta(\text{\textcolor{minecolortau}{$\btau_t$}}|\bs_t, \bz)$, in the offline RL setting. We first lay out a general formulation for such an objective by combining Equations~\ref{eq:model-based-offline-rl} and~\ref{eq:elbo-divserse-skills}: (next, we will mark \textcolor{minecolortau}{$\btau_t$} with \textcolor{minecolortau}{pink} to emphasize that \textcolor{minecolortau}{$\btau_t$} is the output of our learning policy $\pi_\theta(\textcolor{minecolortau}{\btau_t}|\bs_t,\bz)$, different from the offline  $\btau_t \sim \pi_\mathcal{D}(\btau_t)$ used to train the diffusion prior $\psi$ as in Equation~\ref{eq:didi-prior}.) 
\begin{align}
\max_{\pi_\theta, q_\phi} \  \mathbb{E}_{\bz, \bs_t, \textcolor{minecolortau}{\btau_t}} &[ 
\underbrace{\log q_\phi(\bz|\textcolor{minecolortau}{\btau_t}) - \log p(\bz)}_{\text{emergence of diversity}} + {R(\textcolor{minecolortau}{\btau_t})} + \nonumber \\ 
& \underbrace{\log \pi_\mathcal{D}(\textcolor{minecolortau}{\btau_t}) - \log \pi_\theta(\textcolor{minecolortau}{\btau_t})}_{\text{offline regularization}} ], \label{eq:didi-general-formulation}
\end{align} 
where the expectation $\mathbb{E}_{\bz, \bs_t, \textcolor{minecolortau}{\btau_t}} := \mathbb{E}_{p(\bz),\pi_{\mathcal{D}}(\bs_t)\pi_\theta(\textcolor{minecolortau}{\btau_t}|\bs_t,\bz)}$. 

\textbf{Learning diffusion priors.} For the offline regularization term in Equation~\ref{eq:didi-general-formulation}, we first approximate the offline data distribution $\pi_{\mathcal{D}}(\cdot)$ with a forward KL objective (be similar in spirit to \citet{fujimoto2021minimalist}) that learns a unconditional\footnote{The corresponding conditioned diffusion is $\pi_\psi(\btau^{n-1}_t|\btau^n_t, \bz)$ that can not be trained directly because the ``label" $q_\phi(\bz|\btau_t^n)$ is not accessible in the offline RL setting, as described in Section~\ref{sec:planning_with_diffusion_models}.} diffusion generative model $\pi_\psi(\btau^{n-1}_t |\btau^n_t)$ by maximizing:
\begin{align}\label{eq:didi-prior}
\max_{\pi_\psi} \ \mathbb{E}_{\btau_t \sim \pi_\mathcal{D}(\btau_t)}\left[ \log \pi_\psi(\btau_t) \right],
\end{align}
and implement it with Equations~\ref{eq:diffusion-elbo} and~\ref{eq:diffusion-predict-noise}. 
Compared to the popular offline RL methods, such prior learning wrt $\pi_\psi$ is similar in spirit to the typical behavior-policy pre-training step, and approximating with the forward KL also resembles the general BC loss for behavior-policy modeling.

Note that here we denote the diffusion parameters by $\psi$ instead of $\theta$, emphasizing that $\pi_\psi$ is only used during policy training (\textit{as priors}) and it will \textit{not} serve as an inference network at testing. Prior diffusion-based RL~\citep{janner2022planning} conducts inference directly over the diffusion model,  which will hinder the learning of diverse behaviors (discriminator $q_\phi$, Section~\ref{sec:planning_with_diffusion_models}), and encounter expensive inference time at testing, as described in Figure~\ref{fig:diffuser-vs-didi} \textit{top}. 


\textbf{Incorporating priors into skill learning.} 
In essence, given offline data $\btau^0_t := \btau_t \sim p_\mathcal{D}(\btau)$, diffusion model $\pi_\psi$ is trained by introducing \textit{new latent variables} $\btau^{1:N}_t$ (Equation~\ref{eq:diffusion-elbo}) that is specified by a simple diffusion forward (noising) process $q(\btau^{1:N}_t|\btau^{0}_t)$. Here we show how to incorporate such latent variables $\btau^{1:N}_t$ into Equation~\ref{eq:didi-general-formulation} and use the pre-trained diffusion model $\pi_\psi$ to regularized the contextual policy $\pi_\theta(\textcolor{minecolortau}{\btau_t}|\bs_t,\bz)$. 

Assuming trajectory $\textcolor{minecolortau}{\btau_t} \sim \pi_\theta(\textcolor{minecolortau}{\btau_t}|\bs_t,\bz)$ involves latent variables $\bv_t$, \ie $\pi_\theta(\textcolor{minecolortau}{\btau_t}|\bs_t,\bz) = \int_{\bv_t} \pi_\theta(\textcolor{minecolortau}{\btau_t},\bv_t|\bs_t,\bz) \text{d} \bv_t$, we can rewrite Equation~\ref{eq:didi-general-formulation} as: 
\begin{align}
&\mathbb{E}_{\bz, \bs_t, \textcolor{minecolortau}{\btau_t}} \left[ 
{\log q_\phi(\bz|\textcolor{minecolortau}{\btau_t}) - \log p(\bz)} + {R(\textcolor{minecolortau}{\btau_t})} \right] + \label{eq:didi-v}\\ 
&\mathbb{E}_{\bz, \bs_t, \textcolor{minecolortau}{\btau_t}} \left[ 
{\mathbb{E}_{q(\bv|\textcolor{minecolortau}{\btau_t})}} \left[ {\log \pi_{\mathcal{D}}(\textcolor{minecolortau}{\btau_t}, {\bv_t}) - \log \pi_\theta(\textcolor{minecolortau}{\btau_t},{\bv_t}|\bs_t,\bz)} \right]
\right], \nonumber 
\end{align}
where $\pi_\theta(\text{\textcolor{minecolortau}{$\btau_t$}},{\bv_t}|\bs_t,\bz) = \pi_\theta(\textcolor{minecolortau}{\btau_t}|\bs_t,\bz)q({\bv_t}|\textcolor{minecolortau}{\btau_t})$. 
Then, we can specify the output of $\pi_\theta(\textcolor{minecolortau}{\btau_t}|\bs_t,\bz)$ as the instance of $\textcolor{minecolortau}{\btau^0_t}$ that can render a diffusion forward (noising) process $q(\textcolor{minecolortau}{\btau^{1:N}_t} | \textcolor{minecolortau}{\btau^0_t})$ in  diffusion probabilistic models, \eg $\textcolor{minecolortau}{\btau^0_t} := \textcolor{minecolortau}{\btau_t}$, and thus  
we can formulate $q(\bv_t|\textcolor{minecolortau}{\btau_t})$ as the corresponding noising process, \eg setting $\bv_t := \textcolor{minecolortau}{\btau^{1:N}_t}$. 

More importantly, such formulation naturally allows us to replace the distribution $\pi_\mathcal{D}(\textcolor{minecolortau}{\btau^0_t}, \bv_t)$ in Equation~\ref{eq:didi-v} with our pre-trained diffusion prior $\pi_\psi$ (Equation~\ref{eq:didi-prior}): 
\begin{align}
\pi_\mathcal{D}(\textcolor{minecolortau}{\btau_t}, \bv_t) := \pi_\mathcal{D}(\textcolor{minecolortau}{\btau^0_t}, \textcolor{minecolortau}{\btau^{1:N}_t}) \leftarrow  p(\textcolor{minecolortau}{\btau^N_t})\prod_{n=1}^{N}\pi_\psi(\textcolor{minecolortau}{\btau^{n-1}_t}|\textcolor{minecolortau}{\btau^{n}_t}). \nonumber
\end{align} 

Then, analogously to~\citet{ho2020denoising}, we can derive the second expectation in Equation~\ref{eq:didi-v} as 
\begin{align}
&-\mathbb{E}_{\bz, \bs_t, \textcolor{minecolortau}{\btau_t},  \textcolor{minecolortau}{\btau_t^{1:N}}}\left[ 
D_\text{KL}\left( \pi_\theta(\textcolor{minecolortau}{\btau_t}|\bs_t,\bz)  \Vert \pi_\psi(\textcolor{minecolortau}{\btau_t^0}|\textcolor{minecolortau}{\btau^{1}_t}) \right) + 
\textcolor{white}{\sum_{n=1}} \right. \nonumber \\
&\qquad \left.  
\sum_{n>1} D_\text{KL}(q( \textcolor{minecolortau}{\btau^{n-1}_t}| \textcolor{minecolortau}{\btau^n_t}, \textcolor{minecolortau}{\btau_t}) \Vert \pi_\psi( \textcolor{minecolortau}{\btau^{n-1}_t}| \textcolor{minecolortau}{\btau^n_t}))
\right], \label{eq:didi-kl-reg}
\end{align}
where $\mathbb{E}_{\bz, \bs_t, \textcolor{minecolortau}{\btau_t}, \textcolor{minecolortau}{\btau_t^{1:N}}} := \mathbb{E}_{p(\bz),\pi_{\mathcal{D}}(\bs_t)\pi_\theta(\textcolor{minecolortau}{\btau_t}|\bs_t,\bz)q(\textcolor{minecolortau}{\btau^{1:N}_t} | \textcolor{minecolortau}{\btau_t})}$. 

Intuitively, such an objective suggests that we can use a pre-learned diffusion model $\pi_\psi$ to guide/regularize the learning policy $\pi_\theta(\textcolor{minecolortau}{\btau_t}|\bs_t,\bz)$, thus avoiding the out-of-distribution issues in offline RL settings. 
Under the diffusion structure, it not only facilitates learning a single forward policy network (Figure~\ref{fig:diffuser-vs-didi} \textit{bottom}, somewhat like knowledge distillation\footnote{Compared to prior diffusion methods, we learn $\textcolor{minecolortau}{\btau^n_t}$ and set $\pi_\psi$ fixed, while DDPM learns $\pi_\psi$ and samples $\textcolor{minecolortau}{\btau^n_t}$ from fixed dataset.}), but also \textit{retains its flexibility of combining with the other learning objectives}, \eg incorporating the mutual information objective in Equation~\ref{eq:elbo-divserse-skills}.

Combining Equation~\ref{eq:didi-kl-reg} (after reparametrization in Appendix~\ref{app:sec:additional-derivation}) and the first expectation in Equation~\ref{eq:didi-v}, we obtain 
\begin{align}
\mathcal{J}_{\text{DIDI}}(q_\phi, \pi_\theta) := &\mathbb{E}_{\bz, \bs_t, \textcolor{minecolortau}{\btau_t}} \left[ 
{\log q_\phi(\bz|\textcolor{minecolortau}{\btau_t}) - \log p(\bz)} + {R(\textcolor{minecolortau}{\btau_t})} \right] \nonumber \\ 
& - \mathbb{E}_{\bz, \bs_t, \textcolor{minecolortau}{\btau_t}} \left[ 
\mathbb{E}_{n, \textcolor{minecolortau}{\btau^n_t}, \bepsilon} \left[ 
\Vert \bepsilon - \bepsilon_\psi(\textcolor{minecolortau}{\btau^n_t}, n) \Vert^2
\right]
\right], \nonumber 
\end{align}
where $\textcolor{minecolortau}{\btau^n_t} = \sqrt{\bar{\alpha}_n} \textcolor{minecolortau}{\btau_t} + \sqrt{1-\bar{\alpha}_t}\bepsilon$. Then, using the gradient back-propagated through the pre-trained diffusion prior $\psi$, we can learn the contextual policy $\pi_\theta(\textcolor{minecolortau}{\btau_t}|\bs_t,\bz)$ and discriminator $q_\phi(\bz|\textcolor{minecolortau}{\btau^n_t})$ cooperatively.  
To summarize, we provide the pseudo-code of DIDI in Algorithm~\ref{alg:didi}.

\begin{algorithm}[t]
	\caption{Diffusion-Guided Diversity (DIDI)}
	\label{alg:didi}
	\textbf{Require:} offline dataset $\pi_\mathcal{D}(\btau_t)$ and skill distribution $p(\bz)$. 
	Initialize diffusion prior $\pi_\psi$, reward network $R$, skill  discriminator $q_\phi(\bz|\textcolor{minecolortau}{\btau^n_t})$ and contextual policy $\pi_\theta(\textcolor{minecolortau}{\btau^n_t}|\bs_t, \bz)$. 
	
	\begin{algorithmic}[1]
		\STATE Train the diffusion prior $\pi_\psi$ with Equation~\ref{eq:didi-prior}.
		\WHILE{not converged}
		\STATE Sample $\bz \sim p(\bz)$, $\bs_t \sim \pi_\mathcal{D}(\btau_t)$, and $n \sim [0, N]$.
		\STATE Learn $q_\phi(\bz|\textcolor{minecolortau}{\btau^n_t})$ and $\pi_\theta(\textcolor{minecolortau}{\btau^n_t}|\bs_t, \bz)$ with $\mathcal{J}_\text{DIDI}$. 
		\ENDWHILE
	\end{algorithmic}
	\textbf{Return:} contextual policy $\ba_t \sim \pi_\theta(\textcolor{minecolortau}{\btau^n_t}|\bs_t, \bz)$.
\end{algorithm}

\begin{figure*}[ht!]
	\centering 
        \includegraphics[width=0.106\textwidth]{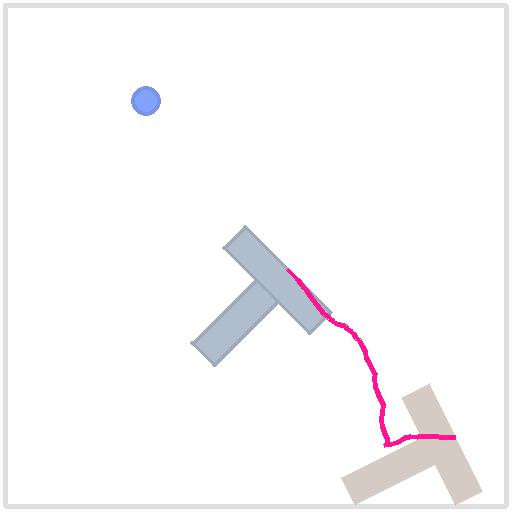}
        \includegraphics[width=0.106\textwidth]{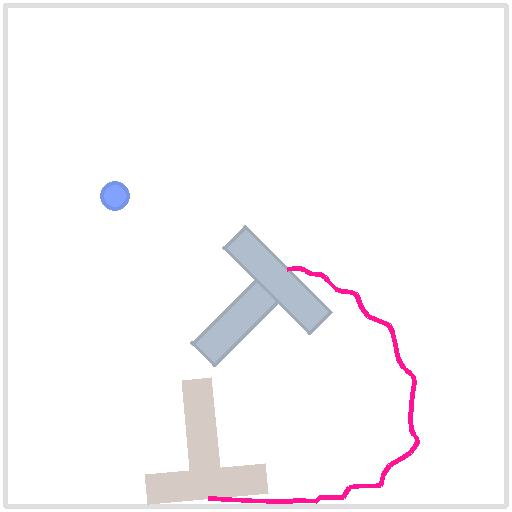}
        \includegraphics[width=0.106\textwidth]{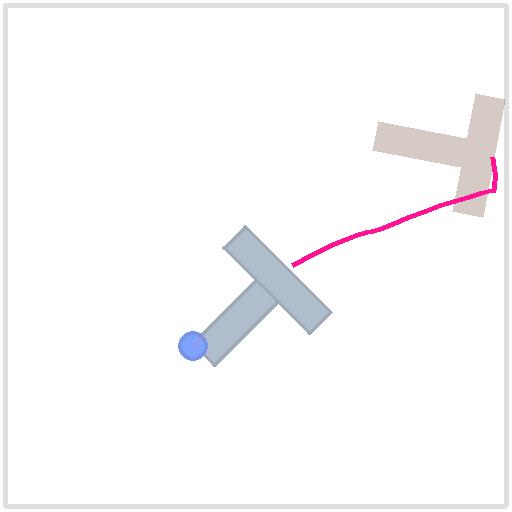}
        \includegraphics[width=0.106\textwidth]{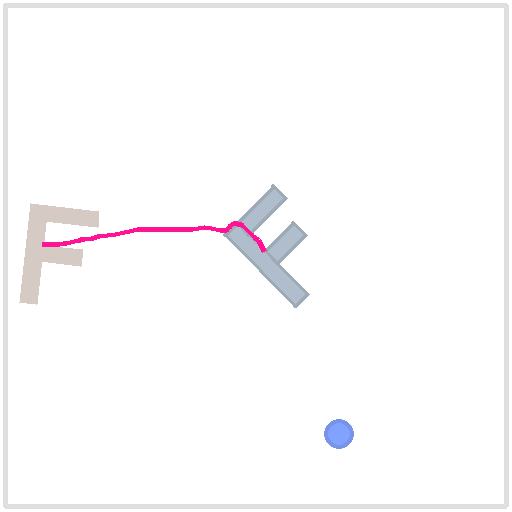}
        \includegraphics[width=0.106\textwidth]{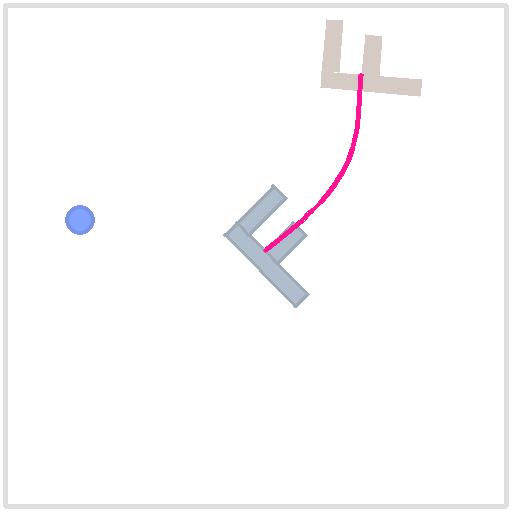}
        \includegraphics[width=0.106\textwidth]{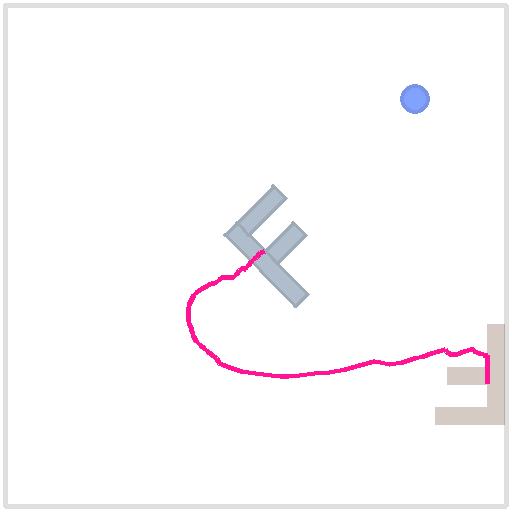}
        \includegraphics[width=0.106\textwidth]{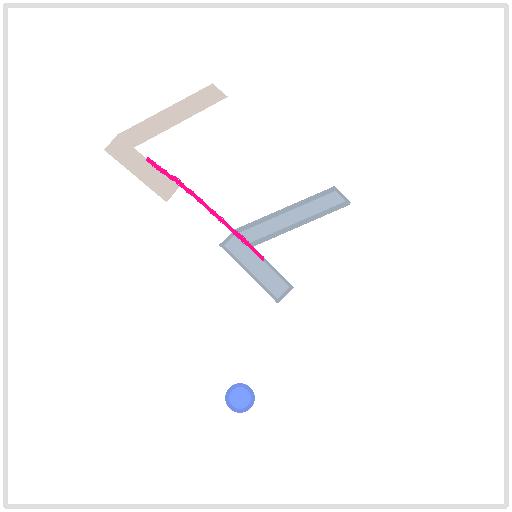}
        \includegraphics[width=0.106\textwidth]{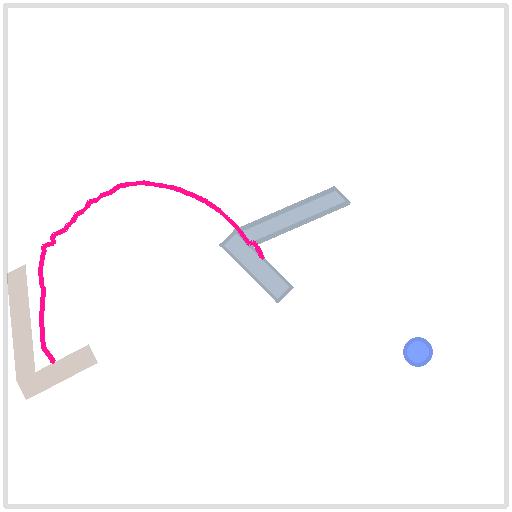}
        \includegraphics[width=0.106\textwidth]{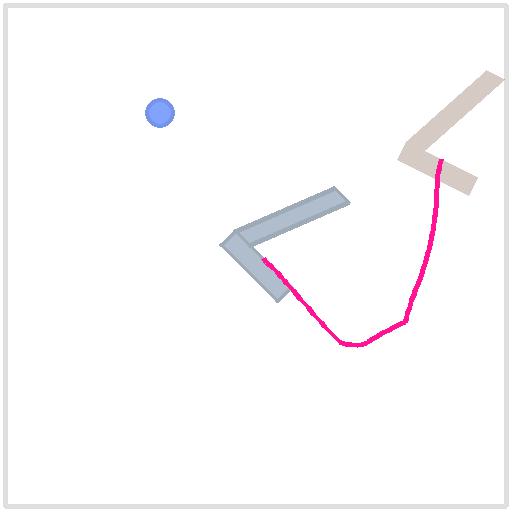}
        \\
        \includegraphics[width=0.1375\textwidth]{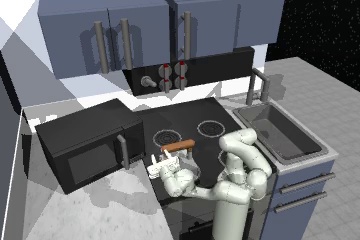}
        \includegraphics[width=0.1375\textwidth]{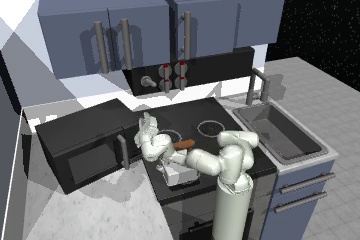}
        \includegraphics[width=0.1375\textwidth]{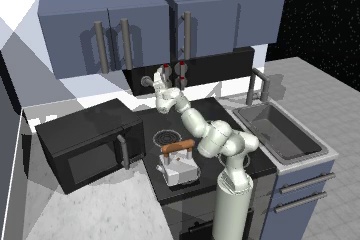}
        \includegraphics[width=0.1375\textwidth]{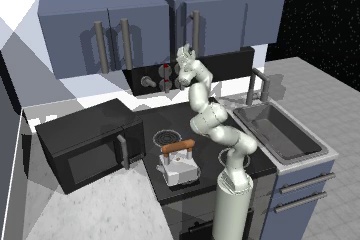}
        \includegraphics[width=0.1375\textwidth]{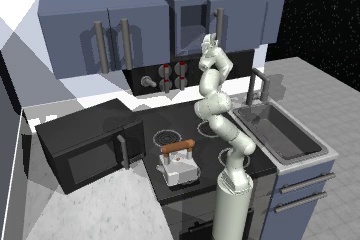}
        \includegraphics[width=0.1375\textwidth]{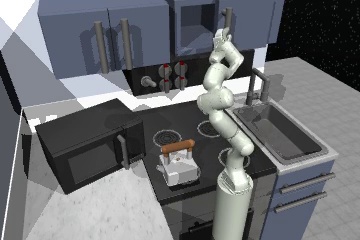}
        \includegraphics[width=0.1375\textwidth]{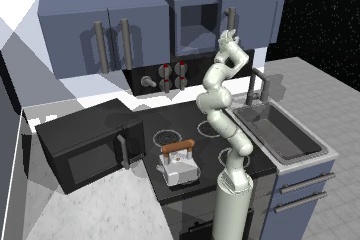}
        \\
        \includegraphics[width=0.1375\textwidth]{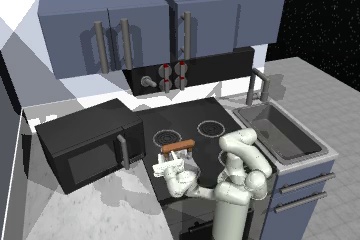}
        \includegraphics[width=0.1375\textwidth]{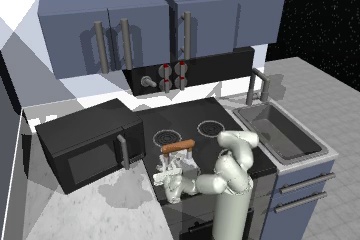}
        \includegraphics[width=0.1375\textwidth]{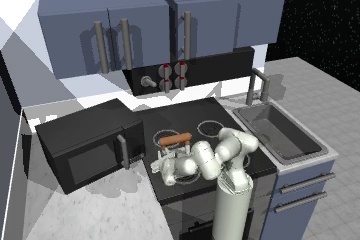}
        \includegraphics[width=0.1375\textwidth]{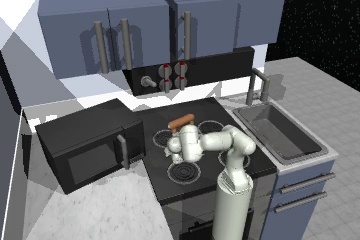}
        \includegraphics[width=0.1375\textwidth]{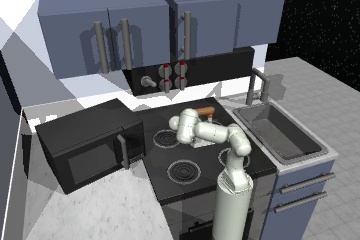}
        \includegraphics[width=0.1375\textwidth]{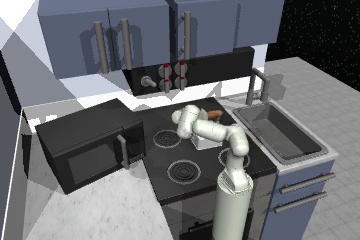}
        \includegraphics[width=0.1375\textwidth]{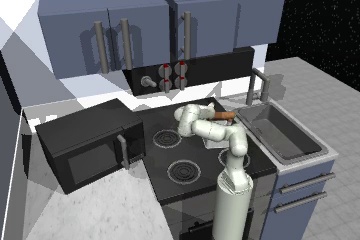}
        \\
        \includegraphics[width=\columnwidth * 2]{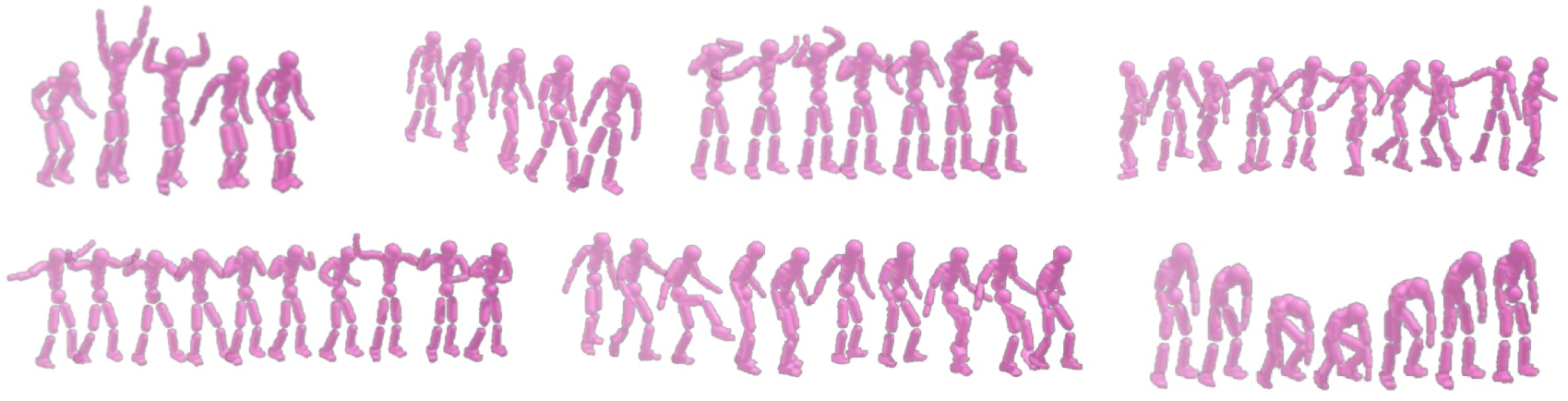}
        \vspace{-7pt}
	\caption{\textbf{Discovered diverse skills in three domains.} We can see that in the Push domain, blocks are pushed to different positions. In the Kitchen domain, the robotic arm executes distinct actions. In the Humanoid domain, the agent exhibits different movements and navigates in different directions (the color progression from light to dark indicates the movement progress of the humanoid).   
	} 
	\label{fig:diverse_skills}
\end{figure*}


\subsection{Variational Auto-Encoders as Priors} 

To gain more insight wrt our objective $\mathcal{J}_\text{DIDI}$, we can also take variational auto-encoders (VAE) as priors for Equations~\ref{eq:didi-general-formulation} and~\ref{eq:didi-v}. Then, we can first learn a VAE prior --- $q_\text{enc}(\bv_t|\btau_t)$ and $p_\text{dec}(\btau_t|\bv_t)$ to approximate the offline data distribution $\pi_\mathcal{D}(\btau_t)$ by maximizing: 
$$\mathbb{E}_{\pi_\mathcal{D}(\btau_t)q_\text{enc}(\bv_t|\btau_t)}\left[\log p_\text{dec}(\btau_t|\bv_t) - \log \frac{q_\text{enc}(\bv_t|\btau_t)}{p(\bv_t)} \right],$$
where $p(\bv_t)$ is a fixed prior. 
Similar to Equation~\ref{eq:didi-kl-reg}, we can derive an alternative form for the second expectation in Equation~\ref{eq:didi-v}: 
\begin{align}
- &\mathbb{E}_{\bz, \bs_t, \textcolor{minecolortau}{\btau_t}, {\bv_t}} \left[ 
D_\text{KL}\left( \pi_\theta(\textcolor{minecolortau}{\btau_t}|\bs_t,\bz)  \Vert q_\text{dec}({\btau_t}|\bv_t) \right) 
\right], \label{eq:didi-kl-reg-vae}
\end{align}
where $\mathbb{E}_{\bz, \bs_t, \textcolor{minecolortau}{\btau_t}, {\bv_t}} := \mathbb{E}_{p(\bz), \pi_\mathcal{D}(\bs_t), \pi_\theta(\textcolor{minecolortau}{\btau_t}|\bs_t, \bz), q_\text{enc}(\bv_t|\textcolor{minecolortau}{\btau_t})}$. 
Intuitively, Equation~\ref{eq:didi-kl-reg-vae} says that we can use the pre-trained decoder $q_\text{dec}({\btau_t}|\bv_t)$ as a target to learn the contextual policy $\pi_\theta(\textcolor{minecolortau}{\btau_t}|\bs_t,\bz)$, but it creates a chicken-and-egg problem that the target \textit{fully} depends on the output of the learning policy itself (\textit{left} sub-diagram, see below):
$$
\left(\smash{{}^{\bz}_{\bs_t}}\right) \stackrel{\pi_\theta}{\longrightarrow} 
\mypoint{tohere}{\textcolor{minecolortau}{\btau_t}} 
\stackrel{q_\text{enc}}{\longrightarrow} 
\mypoint{middlehere}{\bv_t} \stackrel{q_\text{dec}}{\longrightarrow} 
\mypoint{fromhere}{\btau_t} 
\quad \ \ 
\left(\smash{{}^{\bz}_{\bs_t}}\right) \stackrel{\pi_\theta}{\longrightarrow} 
\mypoint{tohere2}{\textcolor{minecolortau}{\btau_t}}  \quad  
\mypoint{middlehere2}{p(\bv_t)} \stackrel{q_\text{dec}}{\longrightarrow} 
\mypoint{fromhere2}{\btau_t} 
$$
{\tikz[overlay, remember picture]{
\draw[-{Stealth[round]}] ([yshift=-2pt]fromhere.south) -- ++(0,-7pt) -| ([yshift=-2pt]tohere.south); 
\node[align=right, below] at ([yshift=-10pt]middlehere.south)  {{\scriptsize As a target for training $\pi_\theta$}}; 
\draw[-] ([yshift=10pt,xshift=10pt]fromhere.east) -- ([yshift=-30pt,xshift=10pt]fromhere.east); 
\draw[-{Stealth[round]}] ([yshift=-2pt]fromhere2.south) -- ++(0,-7pt) -| ([yshift=-2pt]tohere2.south); 
\node[align=right, below] at ([yshift=-10pt]middlehere2.south)  {{\scriptsize As a target for training $\pi_\theta$}}; 
}}

where $q_\text{enc}$ and $q_\text{dec}$ are fixed during the learning of $\pi_\theta$. 

To sidestep this chicken-and-egg problem (leading to unstable training and bad local optima), one can sample $\bv_t$ directly from the ``fixed prior" $p(\bv_t)$ that was used to train the VAE, \ie replacing $q_\text{enc}(\bv_t|\textcolor{minecolortau}{\btau_t})$ with $p(\bv_t)$ as shown in the above diagram (\textit{right}). 
But one has to keep in mind that, essentially, the general objective of $\mathcal{J}_\text{DIDI}$ is trying to cluster offline trajectories into different behavior patterns, which are specified by the contextual variables $\bz$. 
Sampling $\bv_t$ from a ``fixed prior" and using this as the target for learning  $\pi_\theta$ will cause the policy $\pi_\theta(\textcolor{minecolortau}{\btau_t}|\bs_t,\bz)$ to fail to capture diverse behaviors, {because the target} ($q_\text{dec}({\btau_t}|\bv_t), \bv_t \sim p(\bv_t)$) {will not exhibit diversity according to different $\bz$}, and as a result, it will collapse to a single behavior (pursued in normal offline RL). 
Thus, to encourage behavior diversity, we cannot arbitrarily separate the chicken-and-egg connection. 

Going back to Equation~\ref{eq:didi-kl-reg}, we can find that setting $\bv_t=\textcolor{minecolortau}{\btau_t^{1:N}}$ can naturally \textit{yield a balance between the emergence of diversity and the training stability}: 
\textbf{1)} $\bv_t:=\textcolor{minecolortau}{\btau_t^{1:N}}$ is conditioned on the policy's output $\textcolor{minecolortau}{\btau_t}$ according to  $q(\textcolor{minecolortau}{\btau_t^{1:N}}|\textcolor{minecolortau}{\btau_t})$, which thus reserves the chicken-and-egg connection and does not sacrifice the diversity, 
and \textbf{2)} training stability would benefit from large diffusion steps $N$, owing that $\textcolor{minecolortau}{\btau_t^{N}}$ approximates a fixed Gaussian prior $\mathcal{N}(\textbf{0}, \textbf{I})$. 

%
%

\begin{figure*}[t]
        \text{~}\textbf{skill $\bz_{\text{A}}$} 
        \text{~~~~~~~~~~~~~~~~~~~~~~~~~~~~~~~~~~} \textbf{skill $\bz_{\text{B}}$} 
        \text{~~~~~~~~~~~~~~~~~~~~~~~~~~~~~~~~~~~~~~~} \textbf{Stitching: \ \  skill $\bz_{\text{A}}$ $\to$ skill $\bz_{\text{B}}$ $\to$ skill $\bz_{\text{A}}$}
        \\ 
	{ 
        \includegraphics[width=0.45\columnwidth]{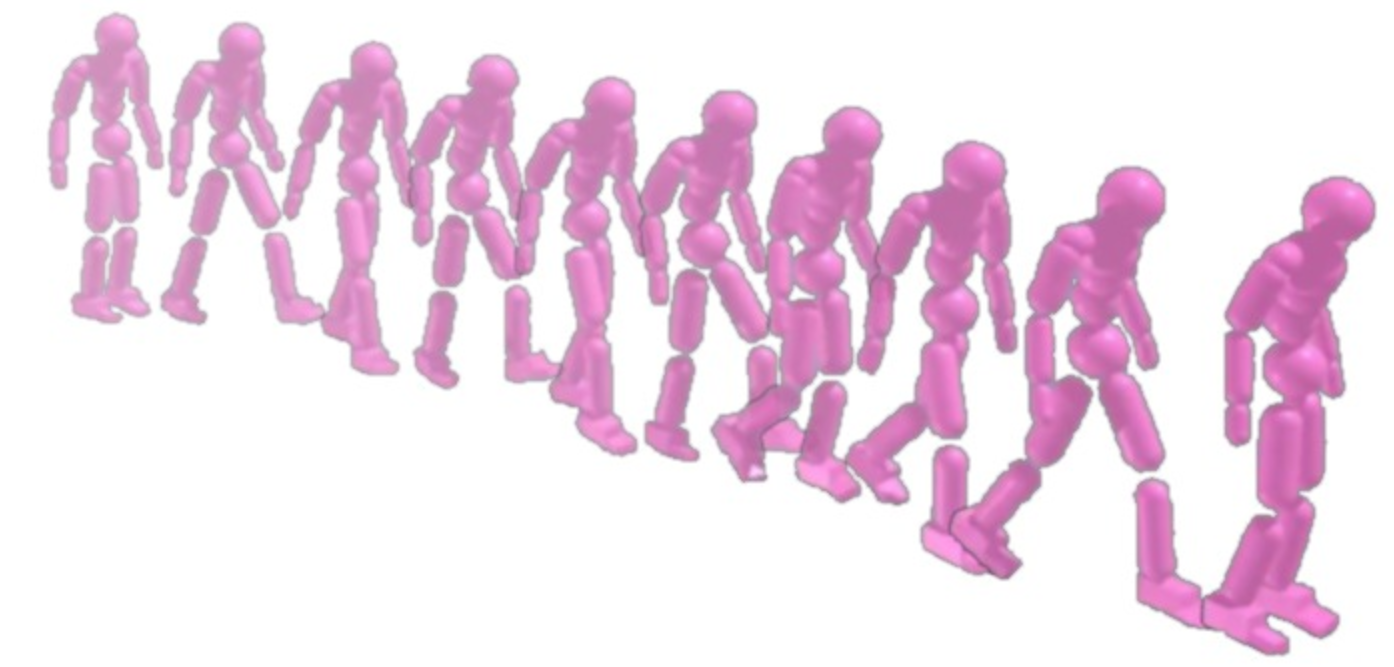} \ \  \ \ 
        \includegraphics[width=0.6\columnwidth]{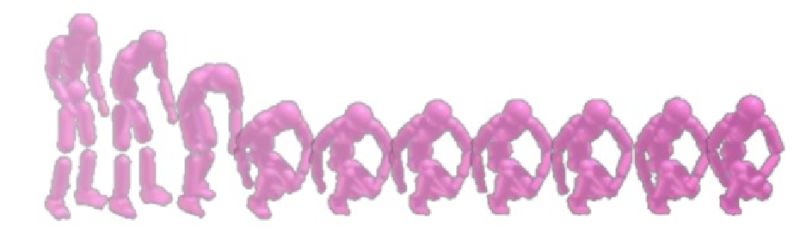} \ \ \ \ 
        \includegraphics[width=0.9\columnwidth]{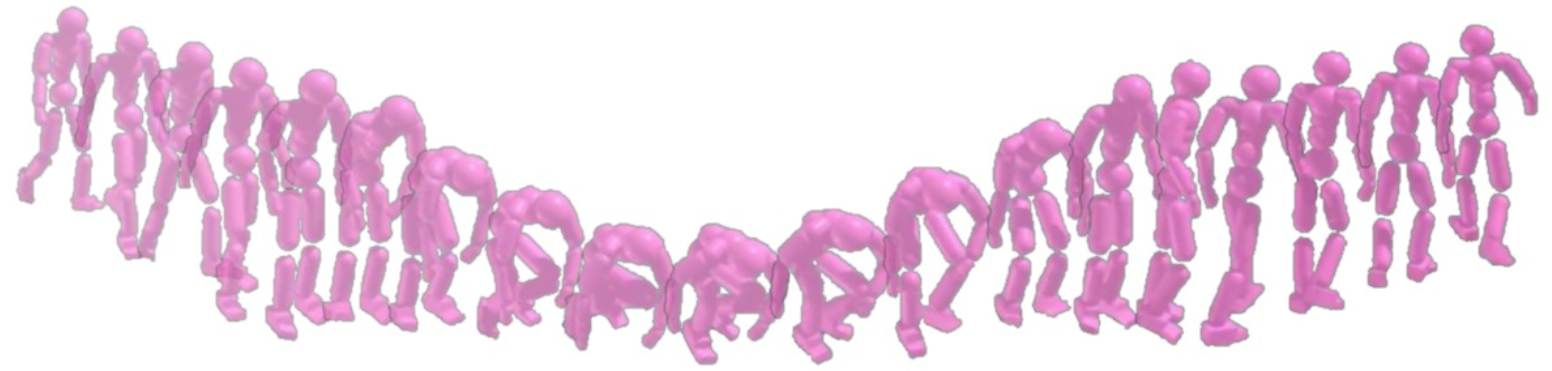}}\\
        \text{~~~~~~~~~~}``walking" 
        \text{~~~~~~~~~~~~~~~~~~~~~~~~~~~~} ``crouching" 
        \text{~~~~~~~~~~~~~~~~~~~~~~~~~~~~~~~~~~~~~~} ``walking" $\to$ ``crouching" $\to$ ``walking"
        \\
        {
        \includegraphics[width=0.45\columnwidth]{figures/skill_sitching_walking_forward.jpg} \ \  \ \ 
        \includegraphics[width=0.6\columnwidth]{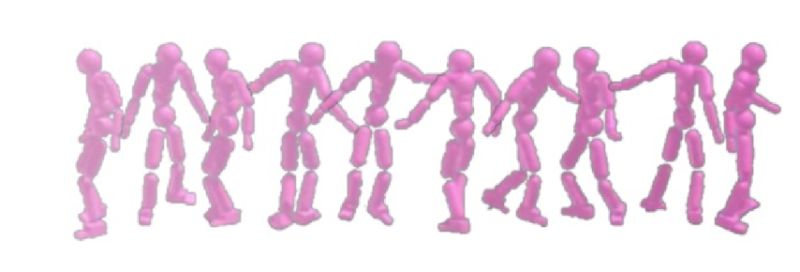} \ \ \ \ 
        \includegraphics[width=0.9\columnwidth]{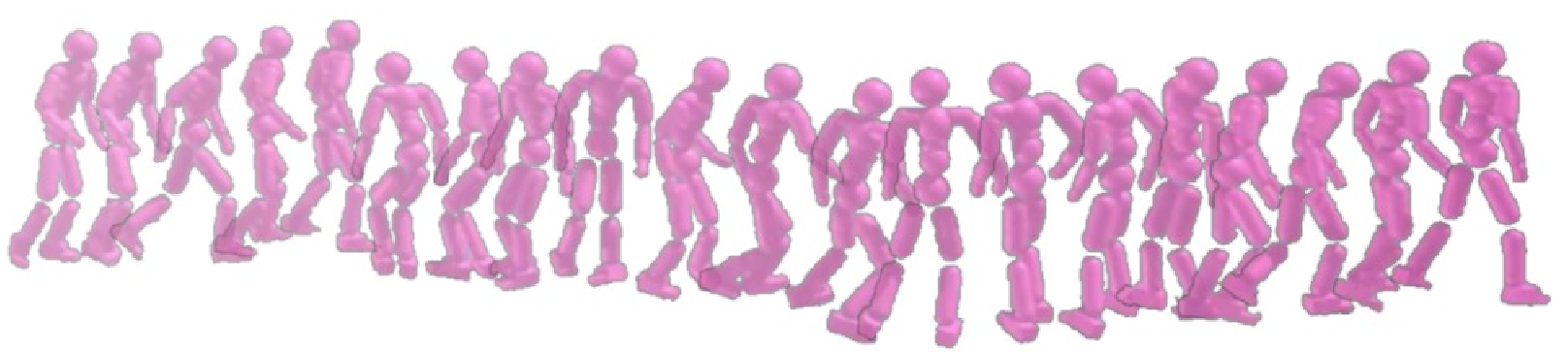}}\\
        \text{~~~~~~~~~~}``walking" 
        \text{~~~~~~~~~~~~~~~~~~~~~~~~~~~~} ``turn round" 
        \text{~~~~~~~~~~~~~~~~~~~~~~~~~~~~~~~~~~~~~~} ``walking" $\to$ ``turn round" $\to$ ``walking"
        \\
        {
        \includegraphics[width=0.45\columnwidth]{figures/skill_sitching_walking_forward.jpg} \ \ \ \ \ \   \ \ 
        \includegraphics[width=0.50\columnwidth] {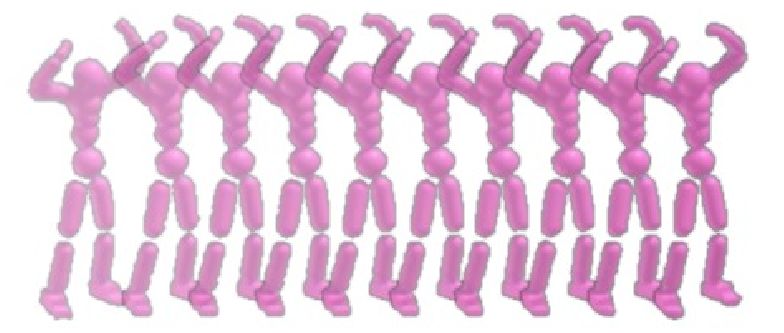} \ \ \ \ \ \ \ \   \ \ 
        \includegraphics[width=0.9\columnwidth]{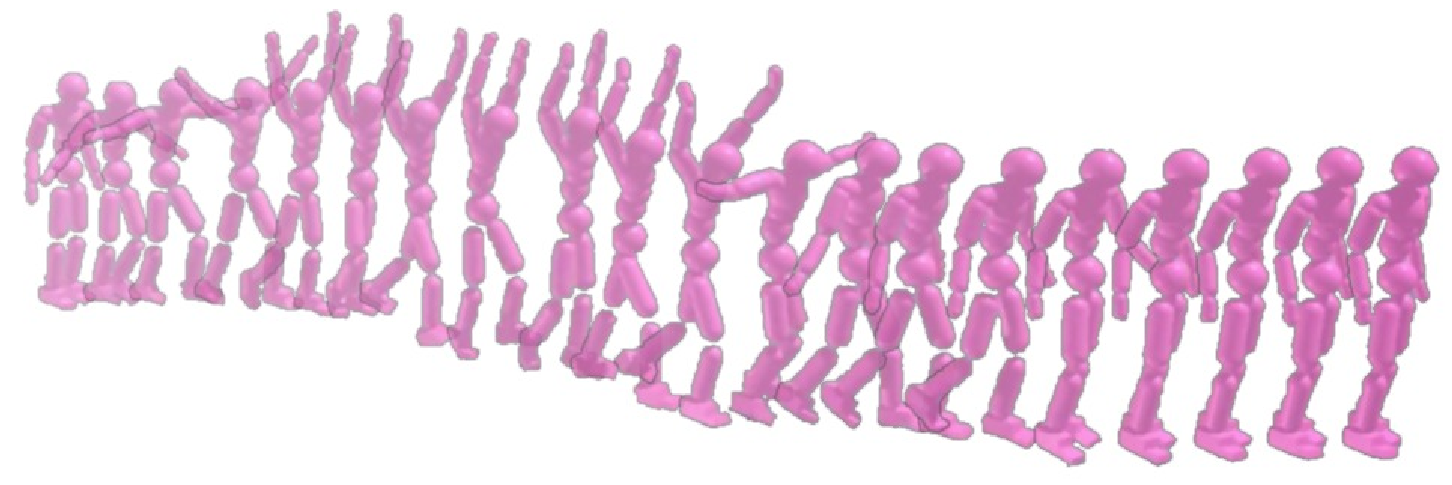}}\\
        \text{~~~~~~~~~~}``walking" 
        \text{~~~~~~~~~~~~~~~~~~~~~~~~~~~~} ``hands up" 
        \text{~~~~~~~~~~~~~~~~~~~~~~~~~~~~~~~~~~~~~~~~} ``walking" $\to$ ``hands up" $\to$ ``walking"
	\caption{\textbf{Skill stitching:} 
        (\textit{1st row}) ``walking" $\to$ ``crouching" $\to$ ``walking", 
        (\textit{2nd row}) ``walking" $\to$ ``turn round" $\to$ ``walking", 
        and (\textit{3rd row}) ``walking" $\to$ ``hands up" $\to$ ``walking". 
        In the diagram, we show (\textit{left}) a ``walking forward" skill and (\textit{middle}) a ``crouching" / ``turn round" / ``hands up" skill, and we find that when the robot is walking forward and we suddenly switch to the "crouching" / ``turn round" / ``hands up" skill, the robot is able to naturally switch the behaviors (\textit{right}). Then, we proceeded to ``walking forward" and the robot could switch back to walking forward. The color progression from light to dark indicates the movement progress of the humanoid. 
        }
	\label{fig:skill_sitching}
\end{figure*}

\section{Experiments}

In our experiments, we aim to answer the following questions\footnote{The code for our implementation is available at  \url{https://github.com/huey0528/icml24didi}.}:
\textbf{1)} Can DIDI discover diverse and discriminative skills from a mixture of (label-free) offline data? 
\textbf{2)} How does DIDI compare to other methods? 
\textbf{3)} As the key point of DIDI is to distill a mixture of behaviors into a low-dimensional skill space, what kind of capabilities can this skill space empower us with? 
\textbf{4)} If an extrinsic reward function is available, can DIDI learn diverse and optimal behaviors from sub-optimal offline data? 
\textbf{5)} What are the benefits of learning a diverse set of behaviors for downstream tasks? 


To answer the above questions, we validate our DIDI in four decision-making domains: Push, Kitchen, Humanoid (as shown in Figure~\ref{fig:diverse_skills}), and D4RL~\citep{fu2020d4rl} tasks. 
The Push task, derived from IBC~\cite{florence2022implicit}, is planning the trajectory to moving a block in a platform with a circular end-effector. 
The Kitchen task~\cite{gupta2019relay} describes the interaction between Franka and seven objects and includes a dataset of 566 human demonstrations. 
The Humanoid task inherits from PHC~\cite{luo2023perpetual}, which focuses on the attainment of high-quality motion imitation. 
The D4RL task, as introduced by ~\citet{fu2020d4rl}, provides a comprehensive suite of benchmark environments designed for offline RL. Specifically, we utilize the Gym-MuJoCo tasks, which involve continuous control environments such as HalfCheetah, Hopper, and Walker2d, to evaluate our DIDI framework's performance (given extrinsic rewards).

\subsection{Emgerence of Behavioral Diversity}

In this section, we investigate the ability of our DIDI approach to discover diverse and discriminative skills from a mixture of label-free offline data. 

Qualitatively, we visualize the discovered behaviors in Figure~\ref{fig:diverse_skills}. 
First, we apply DIDI to the Push domain. 
We can observe that DIDI can learn diverse behaviors that push the block to various positions, demonstrating that DIDI can effectively learn diverse behaviors in this simple task. 
To further evaluate the generality of DIDI, we extend its application to higher-dimensional state and action spaces in the Kitchen and Humanoid tasks. In the Kitchen task, DIDI learns a variety of skills such as  {opening cabinets and picking up kettle}. Similarly, in the Humanoid task, DIDI learns skills like jumping, walking forward, and crouching. These skills cover a wide range of behaviors and showcase the ability of DIDI to discover diverse actions in complex tasks.


\begin{table}
	\centering
        \small 
	\caption{\textbf{Comparison to the behavioral diversity, which qualifies the variance of the motions across all action types.}} %
        \begin{tabular}{lcccccc}
                \toprule
                & Push T & Push F & Push 7 & Kitchen \\
                \midrule
                k-means-DI & 6.8 & 10.4 & 8.1 & 8.7 
                \\
                VAE-DI  & 2.5 & 6.5 &  7.7 & 8.3
                \\
                VAE-DI-fixed  & 1.6 & 7.1  & 5.4 & 9.0
                \\
                DIDI & {7.2} & {12.2} & {10.2} & {9.8}
                \\
                \bottomrule
	\end{tabular}%
	\label{tab:diversity_results}
\vspace{-15pt}
\end{table}

\begin{figure*}[t]
	\begin{center}
        \includegraphics[width=0.106\textwidth]{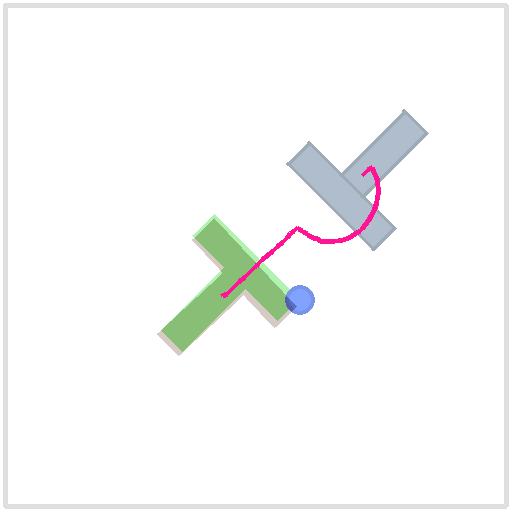}
        \includegraphics[width=0.106\textwidth]{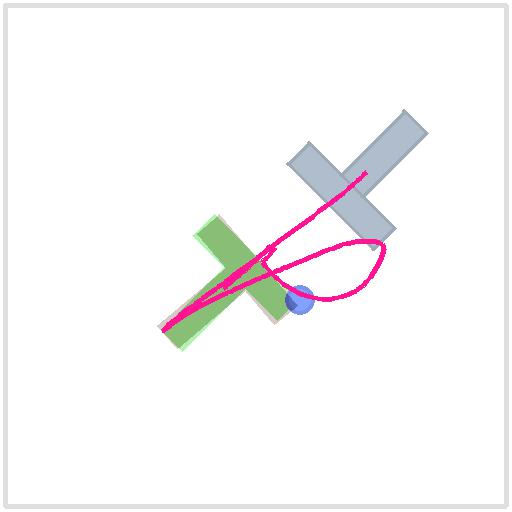}
        \includegraphics[width=0.106\textwidth]{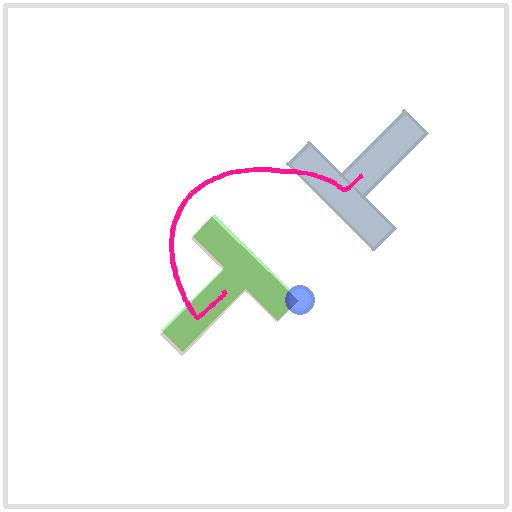}
        \includegraphics[width=0.106\textwidth]{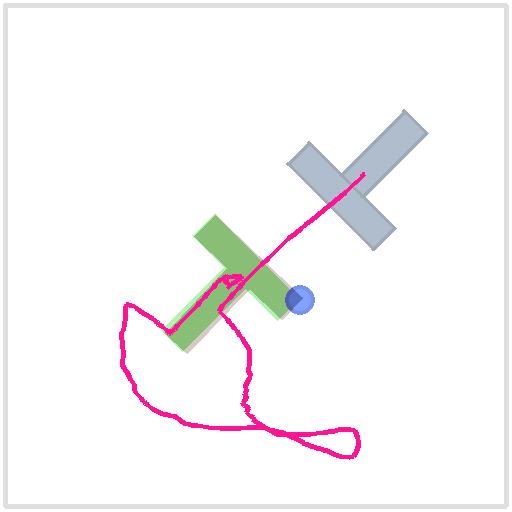}
        \includegraphics[width=0.106\textwidth]{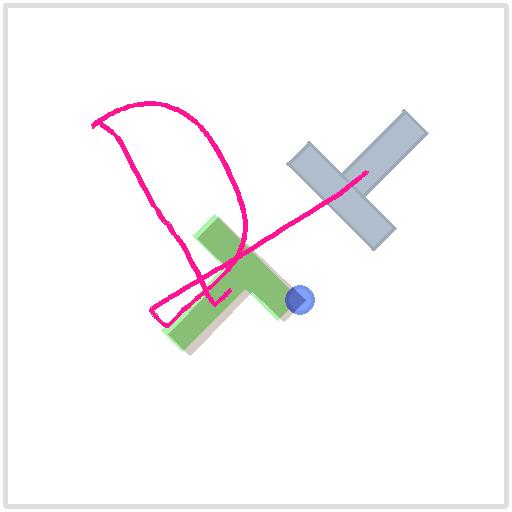}
        \includegraphics[width=0.106\textwidth]{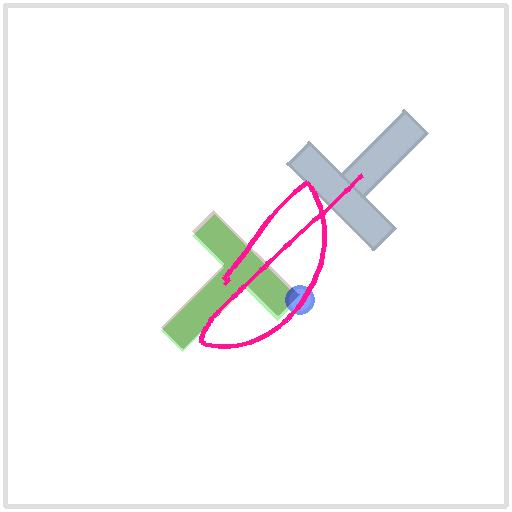}
        \includegraphics[width=0.106\textwidth]{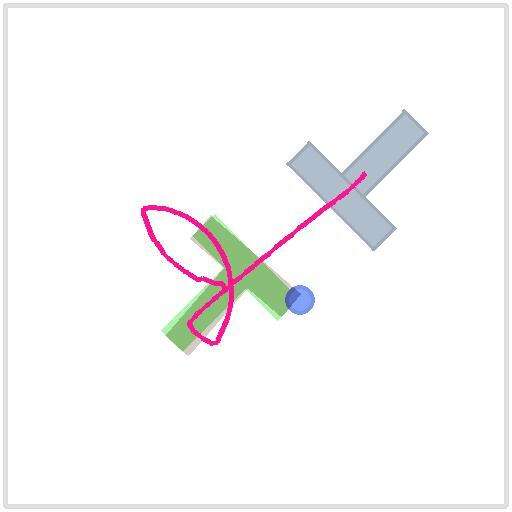}
        \includegraphics[width=0.106\textwidth]{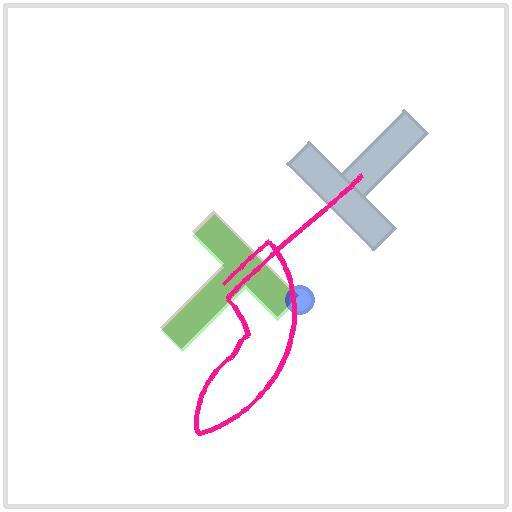}
        \includegraphics[width=0.106\textwidth]{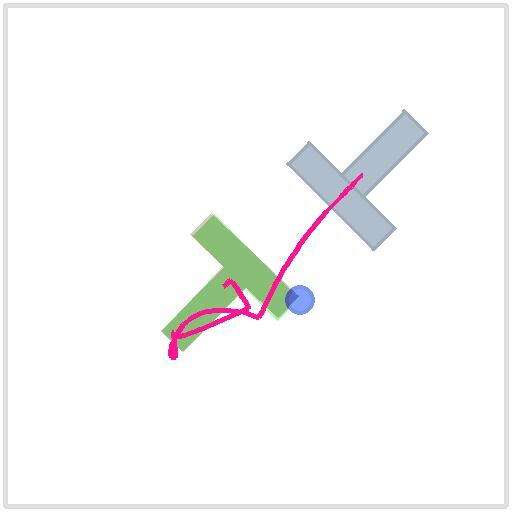}
        \\
        \includegraphics[width=0.106\textwidth]{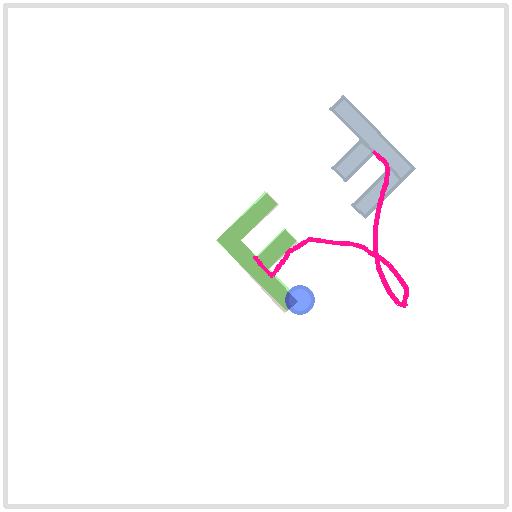}
        \includegraphics[width=0.106\textwidth]{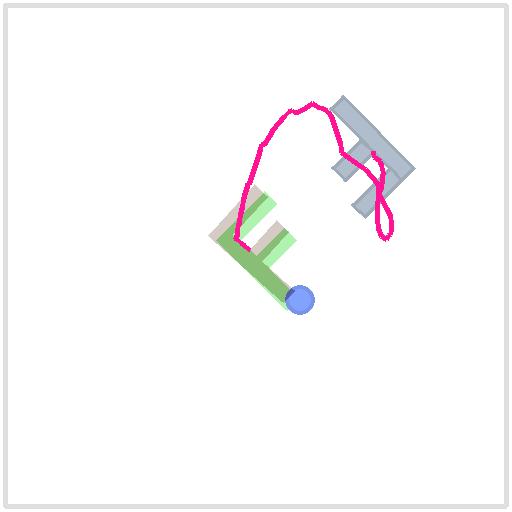}
        \includegraphics[width=0.106\textwidth]{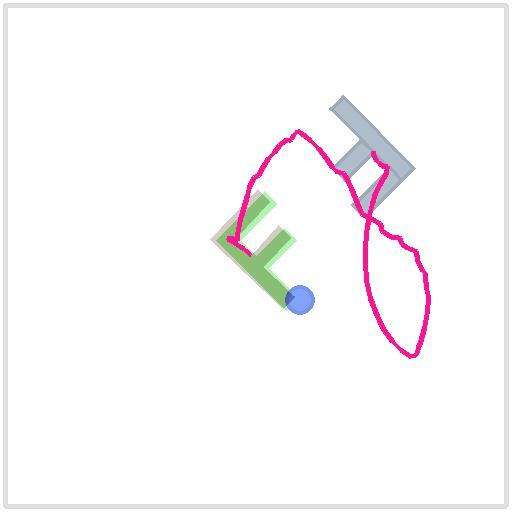}
        \includegraphics[width=0.106\textwidth]{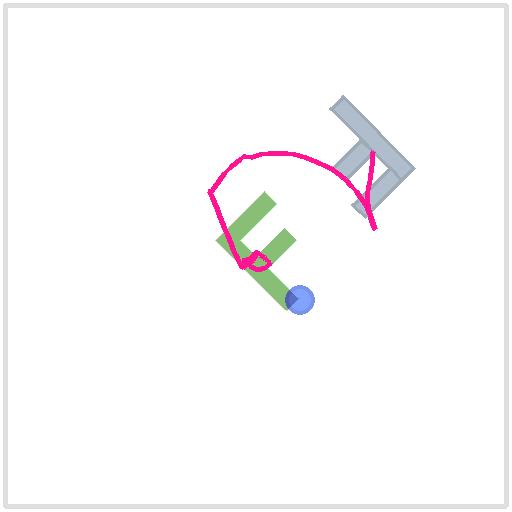}
        \includegraphics[width=0.106\textwidth]{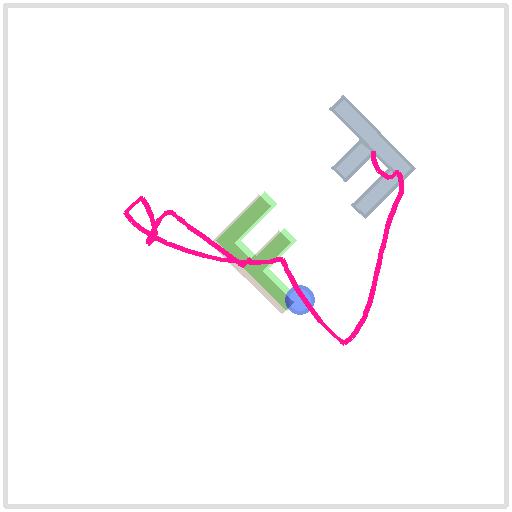}
        \includegraphics[width=0.106\textwidth]{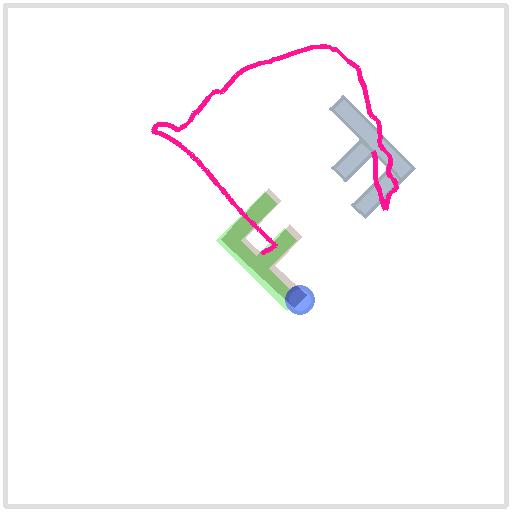}
        \includegraphics[width=0.106\textwidth]{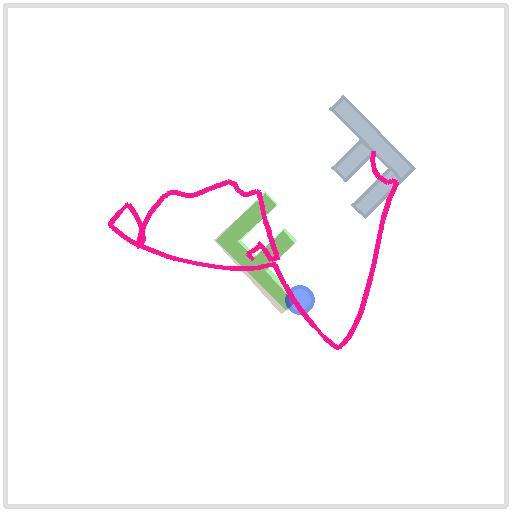}
        \includegraphics[width=0.106\textwidth]{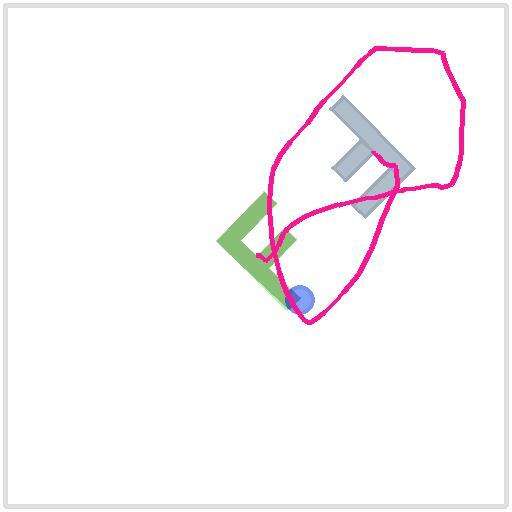}
        \includegraphics[width=0.106\textwidth]{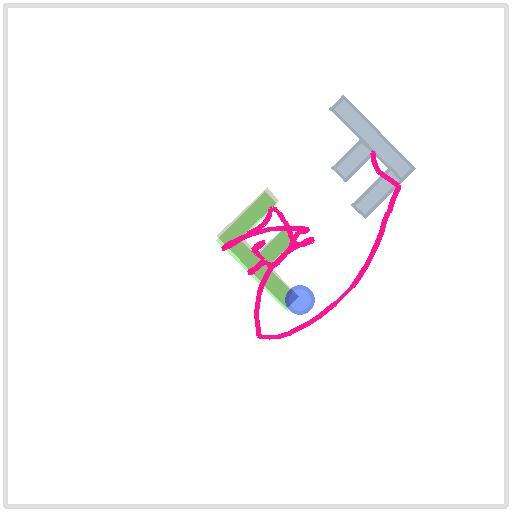}
        \\
        \includegraphics[width=0.106\textwidth]{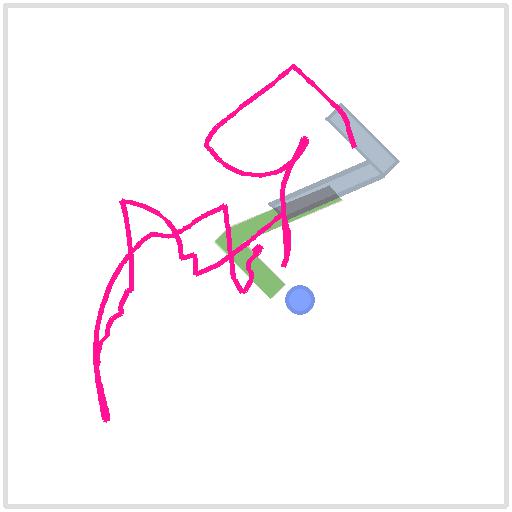}
        \includegraphics[width=0.106\textwidth]{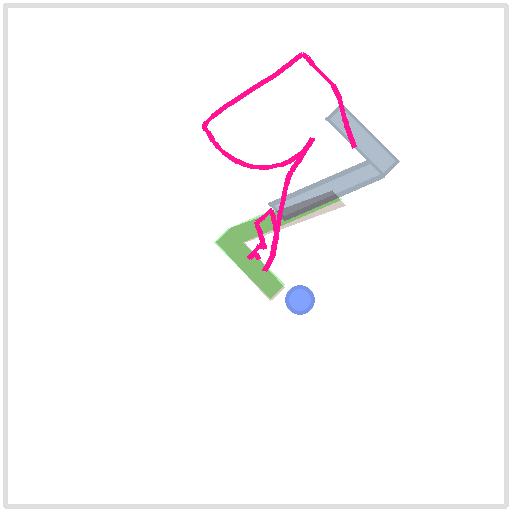}
        \includegraphics[width=0.106\textwidth]{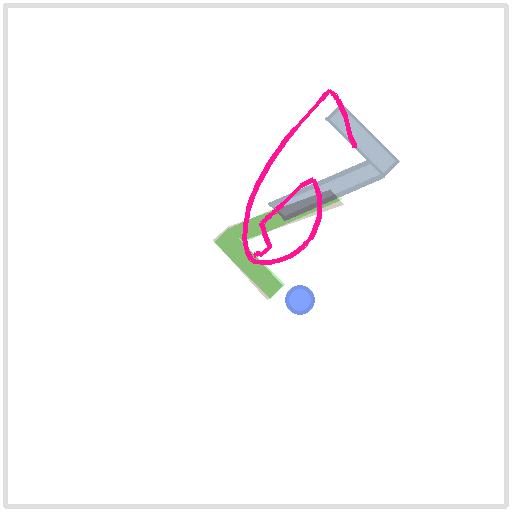}
        \includegraphics[width=0.106\textwidth]{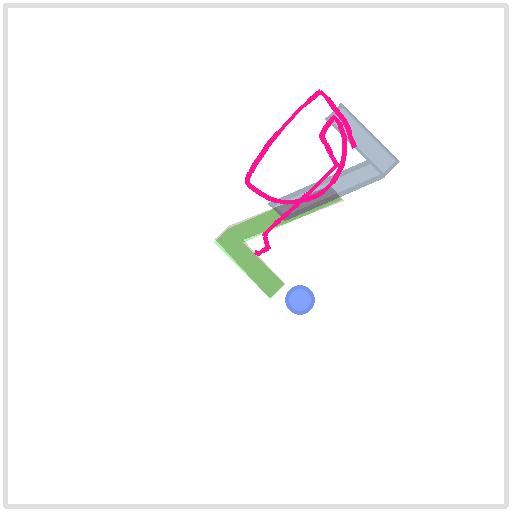}
        \includegraphics[width=0.106\textwidth]{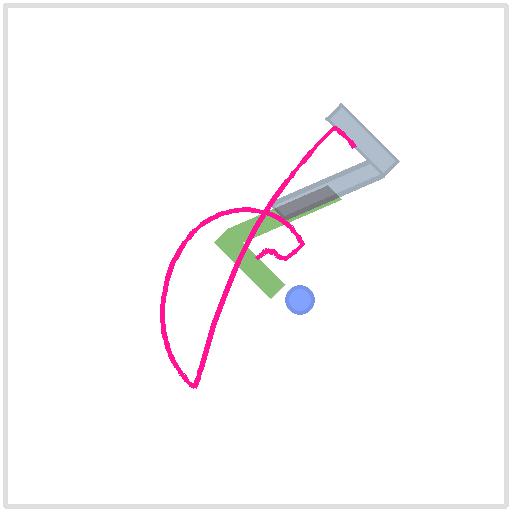}
        \includegraphics[width=0.106\textwidth]{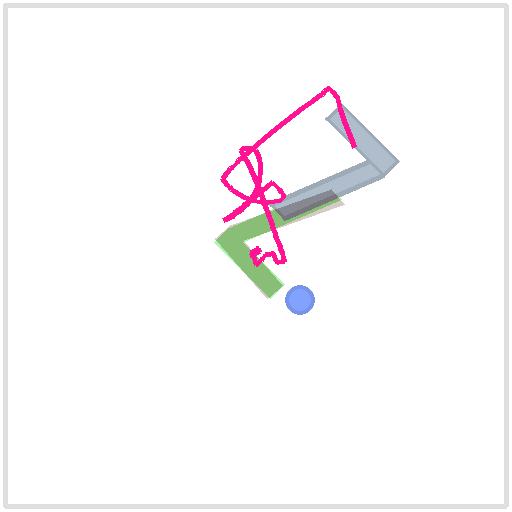}
        \includegraphics[width=0.106\textwidth]{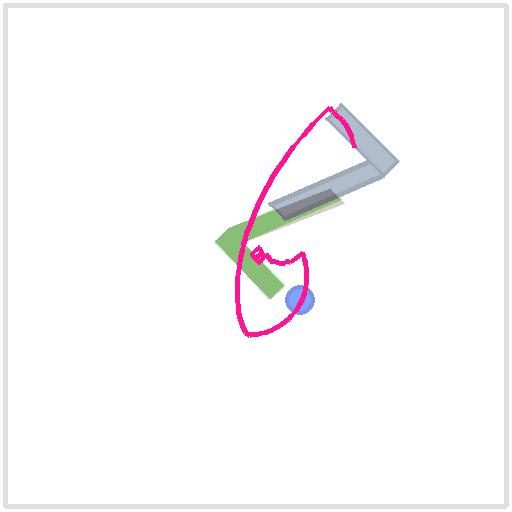}
        \includegraphics[width=0.106\textwidth]{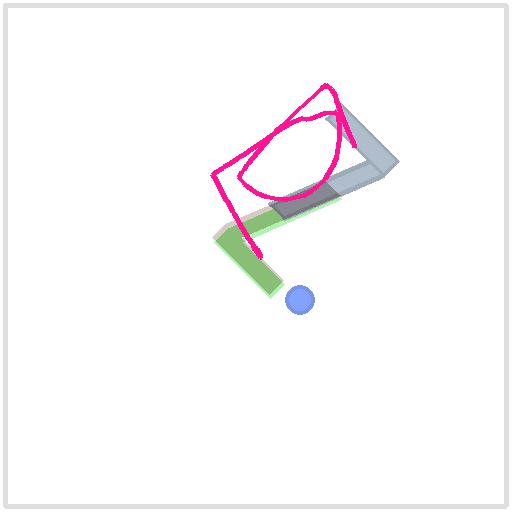}
        \includegraphics[width=0.106\textwidth]{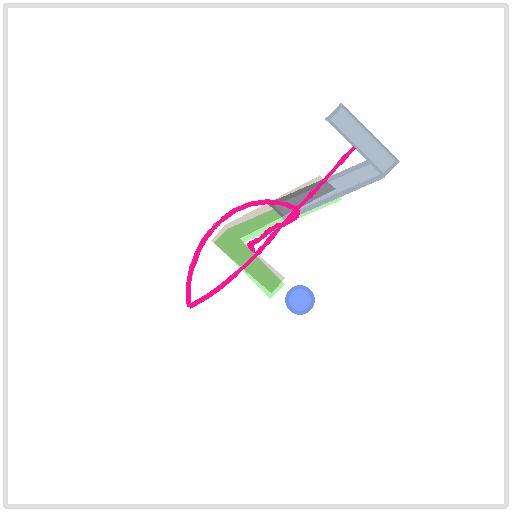}
	\end{center}
        \vspace{-10pt}
	\caption{\textbf{Discovered diverse and optimal skills in the Push domain (with T-shape, F-shape, and 7-shape blocks).} The green block represents the starting point, the gray block represents the target, and the red curve represents the motion trajectory. We can observe that in all tasks, green blocks successfully move to the target positions and display different movement trajectories simultaneously. 
        }
	\label{fig:diversity_and_optimal} 
\end{figure*}

\begin{figure}[t]
	\centering 
	\includegraphics[width=0.70 \columnwidth ]{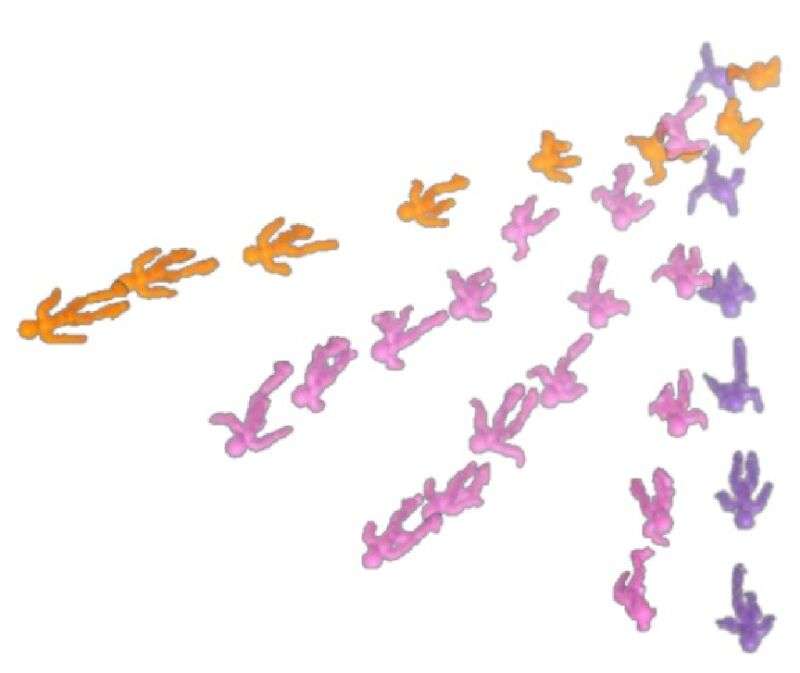}
        \begin{picture}(0,0)
            \put(-160,90){\text{\text{skill $z_{\text{A}}$}}}
            \put(-10,20){\text{\text{skill $z_{\text{B}}$}}}
            \put(-160,120){\text{\textbf{Interpolation: $z_{\text{A}}$ $\to$ $z_{\text{B}}$}}}
        \end{picture}
        \vspace{-10pt}
	\caption{\textbf{Visualization of skill interpolation.} We visualize the top view of skill $z_{\text{A}}$ and skill $z_{\text{B}}$ (Humanoid domain). We can see that by interpolating the skill space, we can obtain the interpolated skills that lie between the movement directions of skills $z_{\text{A}}$ and~$z_{\text{B}}$. 
        }
	\label{fig:skill_interpolation} 
 \vspace{-5pt}
\end{figure}
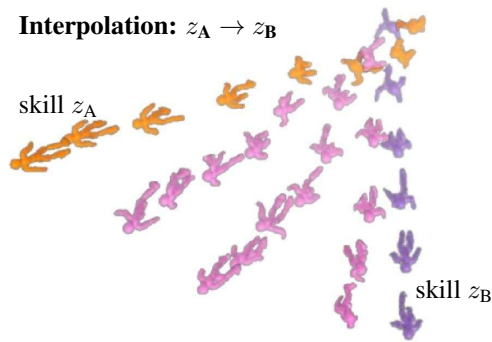

\subsection{Comparison with Alternative Baselines}

To further evaluate the effectiveness of our DIDI approach, we compare it with alternative baselines in terms of skill discovery and performance across Push (including T-shape, F-shape, and 7-shape) and Kitchen domains. 

Specifically, we compare the diversity of skills discovered by DIDI with those obtained by alternative methods: k-means-DI, VAE-DI, and VAE-DI-fixed. 
For k-means-DI, we first use k-means to partition the offline data and then learn a skill policy for each partition; for VAE-DI implementation, we first learn a VAE prior and then use such a prior to guiding the contextual policy, as described in Equation~\ref{eq:didi-kl-reg-vae}; for VAE-DI-fixed, we sample the latent variable $\bv_t$ from the fixed prior $p(\bv_t)$, and use $q_\text{dec}(\bv_t)$ as a target for training~$\pi_\theta$.

We analyze the range of behaviors captured by each method and assess the diversity of the learned skills. 
We use the variance of the motions across all action types as our diversity evaluation metric~\citep{guo2022action2video}. 
We show our diversity scores in Table~\ref{tab:diversity_results}. The results of the experiment prove the effectiveness of DIDI, achieving higher diversity scores. 
This evaluation provides insights into the ability of DIDI to discover diverse and discriminative skills from a mixture of label-free offline data.

\subsection{Generalist Skill Space: Stitching and Interpolation}

Moreover, we also find that our learned skill space tends to be a generalist. For the distilled contextual policy $\pi_\theta(\cdot|\bs_t,\bz)$, their abilities as generalists make them well-suited for skill stitching and interpolation.

\textit{\textbf{Skill stitching}} refers to sequentially combining different skills to perform complex tasks or solve novel problems. As shown in Figure~\ref{fig:skill_sitching}, we can observe that our agents were able to stitch together skills seamlessly: 
(\textit{1st row}) ``walking" $\to$ ``crouching" $\to$ ``walking", 
(\textit{2nd row}) ``walking" $\to$ ``turn round" $\to$ ``walking", 
and (\textit{3rd row}) ``walking" $\to$ ``hands up" $\to$ ``walking". 
And importantly, when we directly command $\pi_\theta$ from skill $z_{\text{A}}$ to skill $z_{\text{B}}$, our learned contextual policy is able to adaptively make adjustments to the actions, without leading to a collapsing behavioral breakdown due to a sudden skill switch. 
Such ability demonstrates the versatility and adaptability of our contextual policy.

\textit{\textbf{Skill interpolation}} is another important aspect of our generalist skill space. It involves smoothly generating behaviors that lie in learned skills. Our agents also exhibit the capability to interpolate skills, allowing them to perform actions and generate new skill behaviors that were not explicitly encountered during training. 
In Figure~\ref{fig:skill_interpolation}, we visualize the top view of skill $z_{\text{A}}$ and skill $z_{\text{B}}$ (Humanoid). We can see that by interpolating the skill space, we can obtain the interpolated skills. 
Thus, such flexibility in skill interpolation showcases the generalization power of our learned skill space.

Furthermore, the generalist nature of our skill space enables the acquisition of new skills through a process of skill refinement and expansion. 
Specifically, this process involves refining existing skills to meet new requirements and expanding the learned behaviors. 
As we will show in the next subsection, by leveraging the existing repertoire of learned skills, our agents can adapt and learn new skills. The ability to continuously learn and expand the skill space will contribute to the overall adaptability of our agents.


\begin{table*}[t]
\centering
\caption{\textbf{Success rates of Push tasks with obstacles after fine-tuning.} The score before the arrow (``$\rightarrow$'') represents fine-tuning a single optimal policy, while the score after the arrow represents fine-tuning our DIDI's behaviors.}
\begin{tabular}{lcccc}
\toprule
\textbf{ } & \textbf{RL (10 ep)} & \textbf{RL (50 ep)} & \textbf{IL (1 demo)} & \textbf{IL (5 demos)} \\
\midrule
Push T & 0.06 $\rightarrow$ 0.40 & 0.18 $\rightarrow$ 0.58 & 0.02 $\rightarrow$ 0.52 & 0.14 $\rightarrow$ 0.82 \\
Push F & 0.04 $\rightarrow$ 0.36 & 0.10 $\rightarrow$ 0.68 & 0.12 $\rightarrow$ 0.48 & 0.24 $\rightarrow$ 0.74 \\
Push 7 & 0.10 $\rightarrow$ 0.44 & 0.12 $\rightarrow$ 0.60 & 0.10 $\rightarrow$ 0.50 & 0.30 $\rightarrow$ 0.88 \\
\bottomrule
\end{tabular}
\label{tab:downstream_results}
\end{table*}


\begin{table}[t]
    \centering
    \caption{\textbf{Quantitative results on D4RL tasks.} Diff.: Diffuser.
    ha: halfcheetah. 
    ho: hopper. 
    wa: walker2d. 
    -m: -medium. 
    -me: -medium-expert. 
    -mr: -medium-replay. 
    } 
    \small 
    \begin{tabular}{lcccccc}
       \toprule
       \multirow{2}{*}{ } & \multirow{2}{*}{CQL} & \multirow{2}{*}{DT} & \multirow{2}{*}{Diff.} & \multicolumn{3}{c}{\textbf{DIDI}} \\
       \cmidrule(lr){5-7}
       & & & & {skill1} & {skill2} & {skill3}  \\
       \midrule
       ha-me & 91.6 & 86.8 & 88.9 & 83.0 & 81.9 & 86.6 \\
       ho-me & 105.4 & 107.6 & 103.3 & 101.2 & 102.1 & 101.9\\
       wa-me & 108.8 & 108.1 & 106.9 & 107.5 & 105.6& 106.9\\
       \midrule
       ha-m & 44.0 & 42.6 & 42.8 & 43.0 & 39.7 & 42.3\\
       ho-m & 58.5 & 67.6 & 74.3 & 70.4 & 74.1 & 72.8\\
       wa-m & 72.5 & 74.0 & 79.6 & 80.3& 78.3 & 79.1 \\
       \midrule
       ha-mr & 45.5 & 36.6 & 37.7 & 34.3 & 33.2 & 34.8\\
       ho-mr & 95.0 & 82.7 & 93.6 & 91.5 & 87.8 & 96.1 \\
       wa-mr & 77.2 & 66.6 &70.6 & 68.9 & 70.3 & 68.1\\
       \midrule
       \multicolumn{1}{l}{Avg.} & 77.6 & 74.7 & 77.5 & 75.6 & 74.8 & 75.7 \\
       \bottomrule
    \end{tabular}
    \label{tab:d4rl}
\vspace{-2pt}
\end{table}

\subsection{Reward-Guided Behavior Generation}

In addition to the generalist nature of our skill space, our DIDI approach also facilitates reward-guided behavior generation. By incorporating an extrinsic reward function, we can leverage DIDI to learn diverse {\textit{and}} optimal behaviors even from sub-optimal offline data. 

In Push domain, we introduce an additional goal-reaching reward function (in Figure~\ref{fig:diversity_and_optimal}, the gray blocks represent the corresponding goal state). Then, we combine both the diversity objective, diffusion regularization, and the goal-reaching reward function to train the contextual policy. We show the learned behaviors in Figure~\ref{fig:diversity_and_optimal}. We can find that all the learned behaviors can reach the goal state, meanwhile providing a rich set of options for the agent to explore and~adapt.

Furthermore, we conduct experiments on D4RL tasks, as shown in Table~\ref{tab:d4rl}. The results indicate that our approach, guided by extrinsic rewards, can learn diverse and optimal skills that achieve comparable performance to standard offline RL methods like CQL and Decision Transformer. 

In Figures~\ref{appendix:fig:walker2d},~\ref{appendix:fig:hopper} and~\ref{appendix:fig:halfcheetah} (appendix), we visualize the learned skills by DIDI, VAE-DI, and Diffuser on D4RL tasks. The visualizations show that the skills learned by DIDI exhibit a high level of diversity compared to those learned by VAE-DI and Diffuser. These results provide further quantitative comparisons, demonstrating that our approach achieves high levels of diversity while maintaining competitive performance with the guidance of extrinsic rewards.

\subsection{Diverse Behaviors for Downstream Tasks}

To demonstrate the practical benefits (of diverse behaviors) on downstream tasks, we conduct tests in two types of downstream settings: fine-tuning in a downstream sparse-reward RL setting and fine-tuning in the few-shot imitation learning (IL) setting. In three Push tasks (Push T, Push F, and Push 7), we randomly introduced obstacles into the environment that could block the robot's path to push the blocks. In this way, we randomly set up 50 downstream tasks (\ie 50 random obstacle/goal configurations).

We fine-tune the learned (single) optimal policy parameters (standard offline RL) for both downstream sparse-reward RL and few-shot imitation learning settings. However, we only fine-tuned the skill embeddings (\ie contextual variables $z$) in our DIDI method, keeping the contextual policy parameters fixed. 
Our experimental results are shown in Table~\ref{tab:downstream_results}, where the online fine-tuning setting provides results for 10 and 50 interactions (episodes) with the environment, and the few-shot imitation fine-tuning setting provides results for 1 and 5 demonstrations. 
In the table, the score before the arrow (``$\rightarrow$'') in each experimental result represents the (fine-tuned) single optimal policy, while the score after the arrow (``$\rightarrow$'') represents the (fine-tuned) DIDI's policies, where the score indicates the success rate of Push tasks (with obstacles). 
We can see that our DIDI method outperforms the baseline (a single policy) in all tested downstream tasks, demonstrating the benefits of our approach (learning diverse behaviors) in enhancing performance in downstream tasks.

\section{Conclusion}

In this paper, we propose a novel approach called DIDI for offline behavioral generation, aiming to learn diverse behaviors from a mixture of label-free offline data. To control the behavioral generation, we introduce a contextual policy that can be commanded to produce specific behaviors. We then use a diffusion probabilistic model as a prior to guide the learning of the contextual policy.
We also compare our approach with the use of variational auto-encoders (VAEs) as priors. We find that our DIDI method better balances diversity (reserving the chicken-and-egg connection) and training stability (approximating a fixed Gaussian prior).

Experimental results in four decision-making domains 
demonstrate the effectiveness of DIDI in discovering diverse and discriminative skills, and generating optimal behaviors from sub-optimal offline data. We also show that DIDI can be used to stitch skills and interpolate between skills, which demonstrates the generalist nature of the learned skill space. 
\section*{Acknowledgements}

The authors would like to express their gratitude to the anonymous reviewers for their valuable comments and suggestions, which have greatly improved the quality of this work.
This work was supported by the National Science and Technology Innovation 2030 - Major Project (Grant No. 2022ZD0208800), and NSFC General Program (Grant No. 62176215).

\section*{Impact Statement}

The potential broader impact of our work lies in several aspects. Firstly, DIDI addresses the challenge of learning from suboptimal or noisy offline data without reward labels, which is a common scenario in real-world applications. By enabling agents to learn from pre-collected large datasets, DIDI eliminates the need for time-consuming and costly online exploration. This can significantly reduce the burden on data collection and accelerate the development of RL applications in various domains.

Secondly, DIDI introduces a controllable behavioral generation approach by incorporating a contextual policy that can be commanded to produce specific behaviors. This controllability is particularly important in practical applications where the ability to command the diversity of behaviors is desired. By allowing users to specify desired behaviors, DIDI opens up possibilities for personalized and adaptive agent behavior in areas such as robotics, gaming, and autonomous systems.

Thirdly, DIDI contributes to the advancement of the field of machine learning by proposing a novel combination of diffusion probabilistic models and unsupervised RL. This integration provides a principled framework for incorporating prior knowledge and regularization into the learning process, leading to improved performance and interpretability of the learned policies. This can have implications for various machine learning tasks beyond offline behavioral generation, such as imitation learning, transfer learning, and model-based RL.


\bibliography{example_paper}
\bibliographystyle{icml2024}

\newpage
\appendix
\onecolumn

\section{Additional Derivation}
\label{app:sec:additional-derivation}

Below is a derivation of our objective $$\mathcal{J}_{\text{DIDI}} := 
\mathbb{E}_{\bz, \bs_t, \textcolor{minecolortau}{\btau_t}} \left[ 
{\log q_\phi(\bz|\textcolor{minecolortau}{\btau_t}) - \log p(\bz)} + {R(\textcolor{minecolortau}{\btau_t})} \right] + 
\mathbb{E}_{\bz, \bs_t, \textcolor{minecolortau}{\btau_t}} \left[ 
\mathbb{E}_{n, \textcolor{minecolortau}{\btau^n_t}, \bepsilon} \left[ 
\Vert \bepsilon - \bepsilon_\psi(\textcolor{minecolortau}{\btau^n_t}, n) \Vert^2
\right]
\right],
$$
which incorporating a pre-trained diffusion model $\bepsilon_\psi$ into learning single-forward policy $\pi_\theta(\textcolor{minecolortau}{\btau^n_t}|\bs_t, \bz)$. 

Starting from the objective in Equation~\ref{eq:didi-v} in the main paper,
\begin{align}
&\mathbb{E}_{\bz, \bs_t, \textcolor{minecolortau}{\btau_t}} \left[ 
{\log q_\phi(\bz|\textcolor{minecolortau}{\btau_t}) - \log p(\bz)} + {R(\textcolor{minecolortau}{\btau_t})} \right] + 
\underbrace{\mathbb{E}_{\bz, \bs_t, \textcolor{minecolortau}{\btau_t}} \left[ 
	{\mathbb{E}_{q(\bv|\textcolor{minecolortau}{\btau_t})}} \left[ {\log \pi_{\mathcal{D}}(\textcolor{minecolortau}{\btau_t}, {\bv_t}) - \log \pi_\theta(\textcolor{minecolortau}{\btau_t},{\bv_t}|\bs_t,\bz)} \right]
	\right]}_{\mathcal{J}_\text{Eq-\ref{eq:didi-v}-II}}.  \tag{Equation~\ref{eq:didi-v}}
\end{align}
we can rewrite the second expectation 
\begin{align}
&\ \ \ \  \mathcal{J}_\text{Eq-\ref{eq:didi-v}-II} \nonumber\\
&=\mathbb{E}_{\bz, \bs_t, \textcolor{minecolortau}{\btau_t}} \left[ 
{\mathbb{E}_{q(\bv_t|\textcolor{minecolortau}{\btau_t})}} \left[ {\log \pi_{\mathcal{D}}(\textcolor{minecolortau}{\btau_t}, {\bv_t}) - \log \pi_\theta(\textcolor{minecolortau}{\btau_t},{\bv_t}|\bs_t,\bz)} \right]
\right] \nonumber\\
&=\mathbb{E}_{\bz, \bs_t, \textcolor{minecolortau}{\btau_t^0}} \left[ 
{\mathbb{E}_{q(\textcolor{minecolortau}{\btau^{1:N}_t}|\textcolor{minecolortau}{\btau_t^0})}} \left[ {\log \pi_{\mathcal{D}}(\textcolor{minecolortau}{\btau_t^0}, {\textcolor{minecolortau}{\btau^{1:N}_t}}) - \log \pi_\theta(\textcolor{minecolortau}{\btau_t^0},{\textcolor{minecolortau}{\btau^{1:N}_t}}|\bs_t,\bz)} \right]
\right] \tag{$\textcolor{minecolortau}{\btau_t^0} := \textcolor{minecolortau}{\btau_t}$, $\bv_t := {\textcolor{minecolortau}{\btau^{1:N}_t}}$}\\
&=\mathbb{E}_{\bz, \bs_t, \textcolor{minecolortau}{\btau_t^0}} \left[ 
{\mathbb{E}_{q(\textcolor{minecolortau}{\btau^{1:N}_t}|\textcolor{minecolortau}{\btau_t^0})}} \left[ {\log p(\textcolor{minecolortau}{\btau^N_t})\prod_{n=1}^{N}\pi_\psi(\textcolor{minecolortau}{\btau^{n-1}_t}|\textcolor{minecolortau}{\btau^{n}_t}) - \log \pi_\theta(\textcolor{minecolortau}{\btau_t^0}|\bs_t,\bz)}\prod_{n=1}^{N}q({\textcolor{minecolortau}{\btau^{n}_t}} | {\textcolor{minecolortau}{\btau^{n-1}_t}}) \right]
\right] \tag{$\pi_\mathcal{D} \leftarrow \pi_\psi$}\\
&=\mathbb{E}_{\bz, \bs_t, \textcolor{minecolortau}{\btau_t^0}} 
\left[ 
{\mathbb{E}_{q(\textcolor{minecolortau}{\btau^{1:N}_t}|\textcolor{minecolortau}{\btau_t^0})}} 
\left[ 
\log p(\textcolor{minecolortau}{\btau^N_t}) 
+ \sum_{n=2}^{N} 
\log \frac{\pi_\psi(\textcolor{minecolortau}{\btau^{n-1}_t}|\textcolor{minecolortau}{\btau^{n}_t})}{q({\textcolor{minecolortau}{\btau^{n}_t}} | {\textcolor{minecolortau}{\btau^{n-1}_t}})} 
+ \log \frac{\pi_\psi(\textcolor{minecolortau}{\btau^{0}_t}|\textcolor{minecolortau}{\btau^{1}_t})}{q(\textcolor{minecolortau}{\btau^{0}_t}|\textcolor{minecolortau}{\btau^{1}_t})} 
- \log \pi_\theta(\textcolor{minecolortau}{\btau_t^0}|\bs_t,\bz)  
\right]
\right] \nonumber\\
&=\mathbb{E}_{\bz, \bs_t, \textcolor{minecolortau}{\btau_t^0}} 
\left[ 
{\mathbb{E}_{q(\textcolor{minecolortau}{\btau^{1:N}_t}|\textcolor{minecolortau}{\btau_t^0})}} 
\left[ 
\log p(\textcolor{minecolortau}{\btau^N_t}) 
+ \sum_{n=2}^{N} 
\log \frac{\pi_\psi(\textcolor{minecolortau}{\btau^{n-1}_t}|\textcolor{minecolortau}{\btau^{n}_t})}{q({\textcolor{minecolortau}{\btau^{n-1}_t}} | {\textcolor{minecolortau}{\btau^{n}_t}, \textcolor{minecolortau}{\btau^{0}_t}})} 
\cdot 
\frac{q({\textcolor{minecolortau}{\btau^{n-1}_t}} | \textcolor{minecolortau}{\btau^{0}_t})}{q({\textcolor{minecolortau}{\btau^{n}_t}} | \textcolor{minecolortau}{\btau^{0}_t})}
+ \log \frac{\pi_\psi(\textcolor{minecolortau}{\btau^{0}_t}|\textcolor{minecolortau}{\btau^{1}_t})}{q(\textcolor{minecolortau}{\btau^{0}_t}|\textcolor{minecolortau}{\btau^{1}_t})} 
- \log \pi_\theta(\textcolor{minecolortau}{\btau_t^0}|\bs_t,\bz)  
\right]
\right] \nonumber\\
&=\mathbb{E}_{\bz, \bs_t, \textcolor{minecolortau}{\btau_t^0}} 
\left[ 
{\mathbb{E}_{q(\textcolor{minecolortau}{\btau^{1:N}_t}|\textcolor{minecolortau}{\btau_t^0})}} 
\left[ 
\log \frac{p(\textcolor{minecolortau}{\btau^N_t}) }{q({\textcolor{minecolortau}{\btau^{N}_t}} | \textcolor{minecolortau}{\btau^{0}_t})}
+ \sum_{n=2}^{N} 
\log \frac{\pi_\psi(\textcolor{minecolortau}{\btau^{n-1}_t}|\textcolor{minecolortau}{\btau^{n}_t})}{q({\textcolor{minecolortau}{\btau^{n-1}_t}} | {\textcolor{minecolortau}{\btau^{n}_t}, \textcolor{minecolortau}{\btau^{0}_t}})} 
+ \log {\pi_\psi(\textcolor{minecolortau}{\btau^{0}_t}|\textcolor{minecolortau}{\btau^{1}_t})}
- \log \pi_\theta(\textcolor{minecolortau}{\btau_t^0}|\bs_t,\bz)  
\right]
\right] \nonumber\\
&=\mathbb{E}_{\bz, \bs_t, \textcolor{minecolortau}{\btau_t^0}}  
\left[ 
{\mathbb{E}_{q(\textcolor{minecolortau}{\btau^{1:N}_t}|\textcolor{minecolortau}{\btau_t^0})}} 
\left[ 
- D_\text{KL}\left( q({\textcolor{minecolortau}{\btau^{N}_t}} | \textcolor{minecolortau}{\btau^{0}_t}) \Vert p(\textcolor{minecolortau}{\btau^N_t}) \right)
 - \sum_{n=2}^{N} 
D_\text{KL}\left( q({\textcolor{minecolortau}{\btau^{n-1}_t}} | {\textcolor{minecolortau}{\btau^{n}_t}, \textcolor{minecolortau}{\btau^{0}_t}}) \Vert \pi_\psi(\textcolor{minecolortau}{\btau^{n-1}_t}|\textcolor{minecolortau}{\btau^{n}_t}) \right) \right. \right. \nonumber\\
& \ \ \ \qquad \qquad \qquad \qquad \qquad \qquad \qquad \qquad \qquad \quad 
\left. 
\left. 
\textcolor{white}{\sum_{n=1}} 
- D_\text{KL}\left( \pi_\theta(\textcolor{minecolortau}{\btau_t^0}|\bs_t,\bz)  \Vert \pi_\psi(\textcolor{minecolortau}{\btau^{0}_t}|\textcolor{minecolortau}{\btau^{1}_t}) \right) 
\right] 
\right] \nonumber 
\end{align}
In the implementation, we set $q({\textcolor{minecolortau}{\btau^{1:N}_t}} | \textcolor{minecolortau}{\btau^{0}_t})$ as a Gaussian noising process, thus the first KL term $D_\text{KL}\left( q({\textcolor{minecolortau}{\btau^{N}_t}} | \textcolor{minecolortau}{\btau^{0}_t}) \Vert p(\textcolor{minecolortau}{\btau^N_t}) \right)$ is a constant and can be ignored. 


\section{Additional Results}

\begin{figure*}[ht!]
	\centering 
        \includegraphics[width=0.13\textwidth]{figures/diversity_only_t/1.jpeg}
        \includegraphics[width=0.13\textwidth]{figures/diversity_only_t/2.jpeg}
        \includegraphics[width=0.13\textwidth]{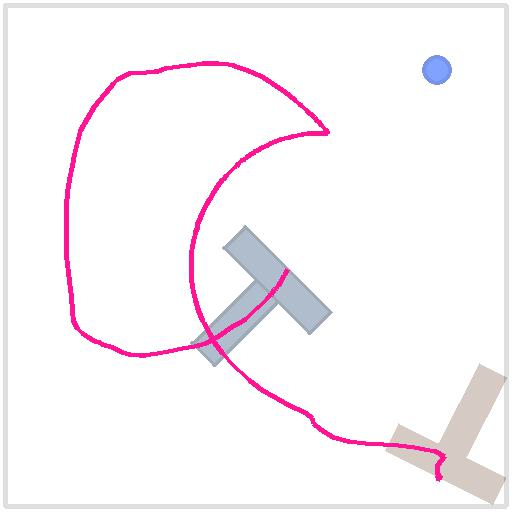}
        \includegraphics[width=0.13\textwidth]{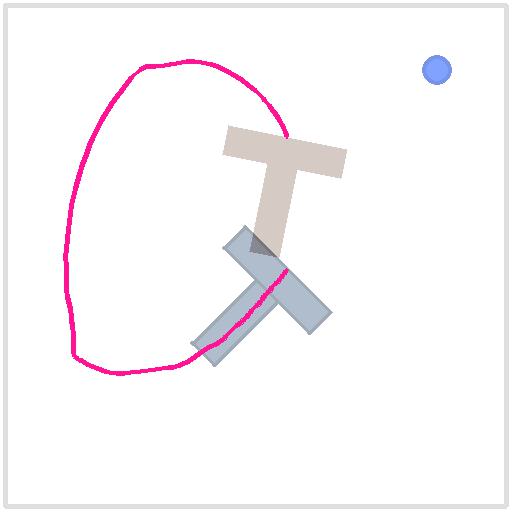}
        \includegraphics[width=0.13\textwidth]{figures/diversity_only_t/5.jpeg}
        \includegraphics[width=0.13\textwidth]{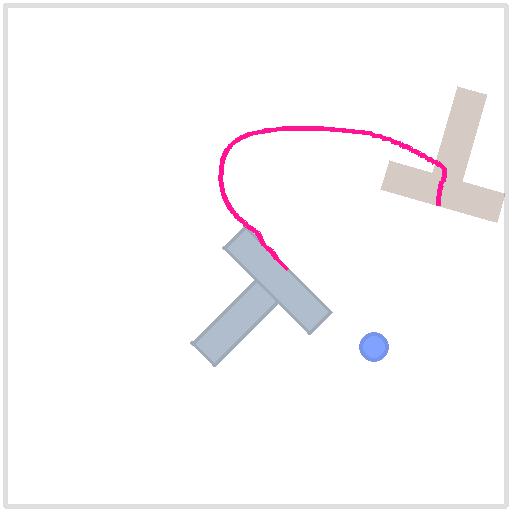}
        \includegraphics[width=0.13\textwidth]{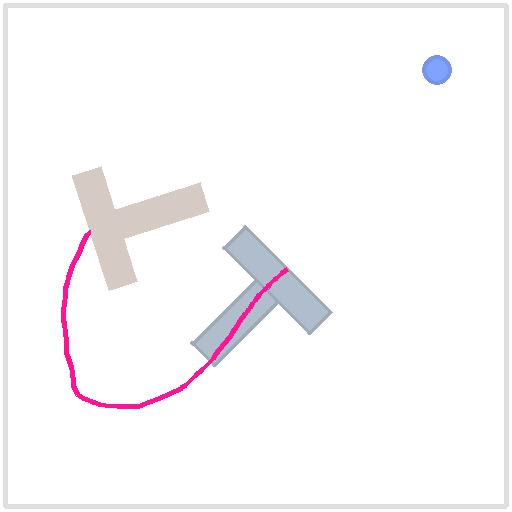}
        \\
        \includegraphics[width=0.13\textwidth]{figures/diversity_only_f/1.jpeg}
        \includegraphics[width=0.13\textwidth]{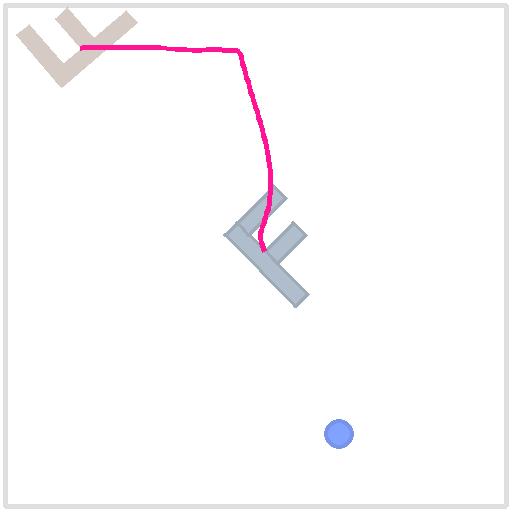}
        \includegraphics[width=0.13\textwidth]{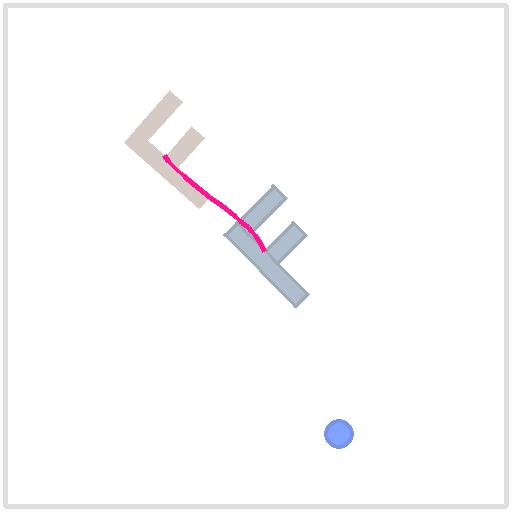}
        \includegraphics[width=0.13\textwidth]{figures/diversity_only_f/4.jpeg}
        \includegraphics[width=0.13\textwidth]{figures/diversity_only_f/5.jpeg}
        \includegraphics[width=0.13\textwidth]{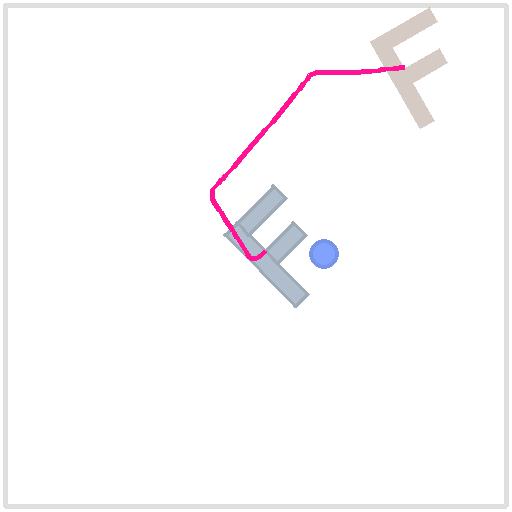}
        \includegraphics[width=0.13\textwidth]{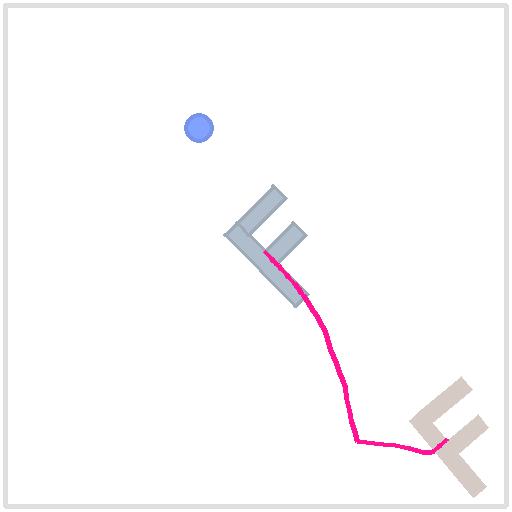}
        \\
        \includegraphics[width=0.13\textwidth]{figures/diversity_only_7/1.jpeg}
        \includegraphics[width=0.13\textwidth]{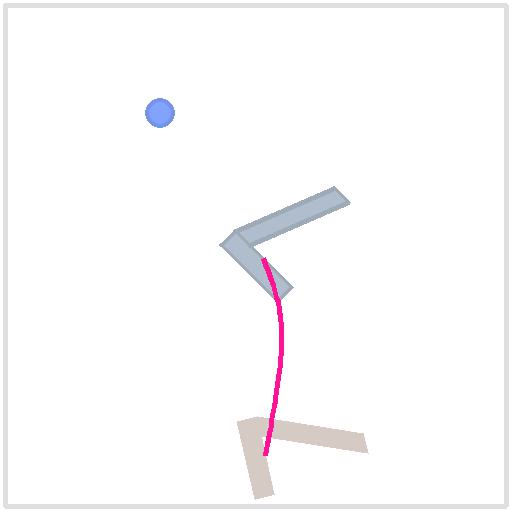}
        \includegraphics[width=0.13\textwidth]{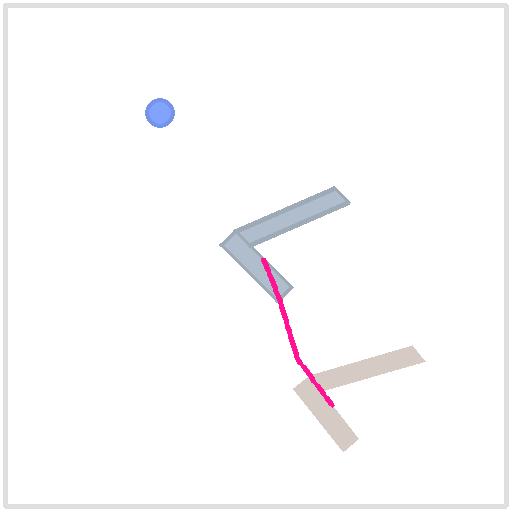}
        \includegraphics[width=0.13\textwidth]{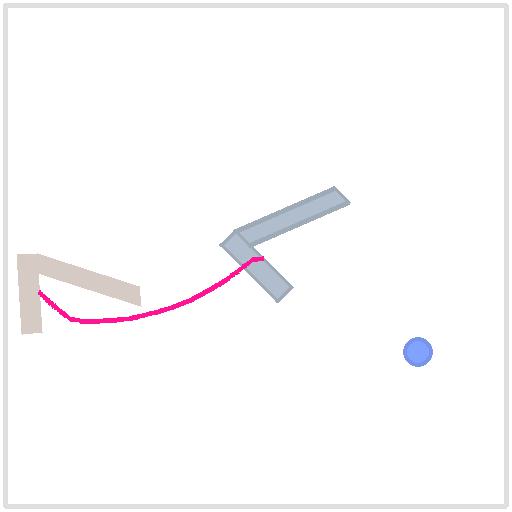}
        \includegraphics[width=0.13\textwidth]{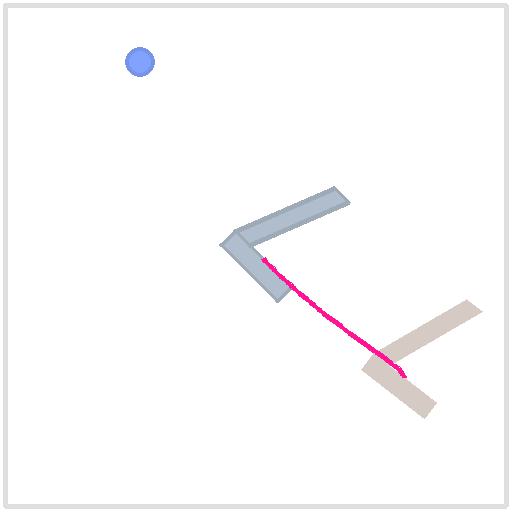}
        \includegraphics[width=0.13\textwidth]{figures/diversity_only_7/6.jpeg}
        \includegraphics[width=0.13\textwidth]{figures/diversity_only_7/7.jpeg}
        \\
        \includegraphics[width=0.130\textwidth]{figures/kitchen/cabinet1.jpg}
        \includegraphics[width=0.130\textwidth]{figures/kitchen/cabinet2.jpg}
        \includegraphics[width=0.130\textwidth]{figures/kitchen/cabinet3.jpg}
        \includegraphics[width=0.130\textwidth]{figures/kitchen/cabinet4.jpg}
        \includegraphics[width=0.130\textwidth]{figures/kitchen/cabinet5.jpg}
        \includegraphics[width=0.130\textwidth]{figures/kitchen/cabinet6.jpg}
        \includegraphics[width=0.130\textwidth]{figures/kitchen/cabinet7.jpg}
        \\
        \includegraphics[width=0.130\textwidth]{figures/kitchen/kettle1.jpg}
        \includegraphics[width=0.130\textwidth]{figures/kitchen/kettle2.jpg}
        \includegraphics[width=0.130\textwidth]{figures/kitchen/kettle3.jpg}
        \includegraphics[width=0.130\textwidth]{figures/kitchen/kettle4.jpg}
        \includegraphics[width=0.130\textwidth]{figures/kitchen/kettle5.jpg}
        \includegraphics[width=0.130\textwidth]{figures/kitchen/kettle6.jpg}
        \includegraphics[width=0.130\textwidth]{figures/kitchen/kettle7.jpg}
        \\
        \includegraphics[width=0.130\textwidth]{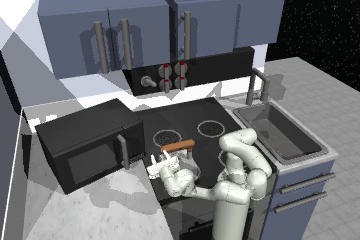}
        \includegraphics[width=0.130\textwidth]{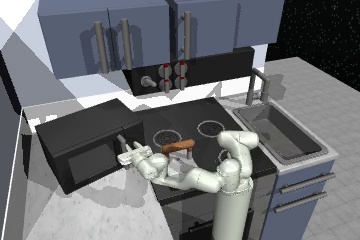}
        \includegraphics[width=0.130\textwidth]{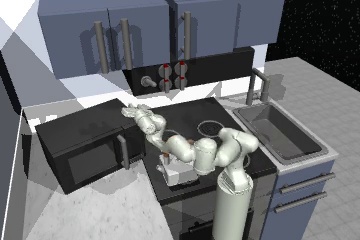}
        \includegraphics[width=0.130\textwidth]{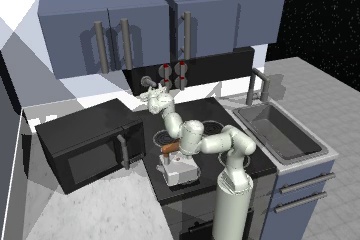}
        \includegraphics[width=0.130\textwidth]{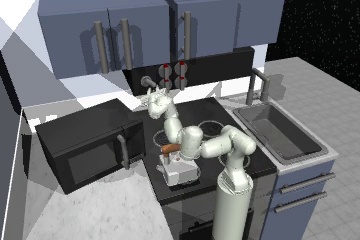}
        \includegraphics[width=0.130\textwidth]{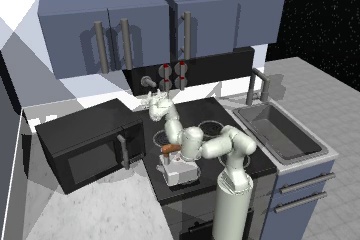}
        \includegraphics[width=0.130\textwidth]{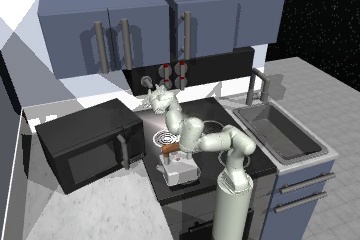}
        \\
        \includegraphics[width=0.130\textwidth]{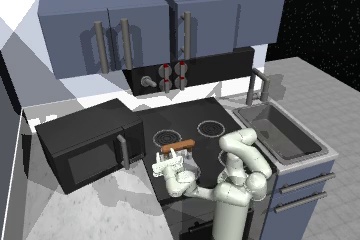}
        \includegraphics[width=0.130\textwidth]{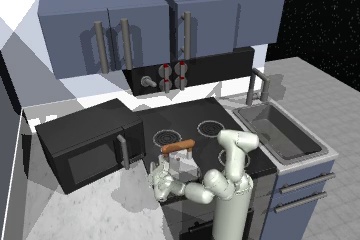}
        \includegraphics[width=0.130\textwidth]{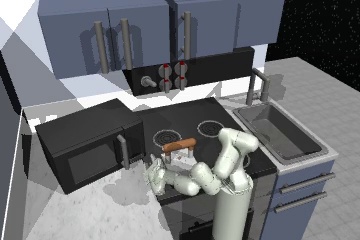}
        \includegraphics[width=0.130\textwidth]{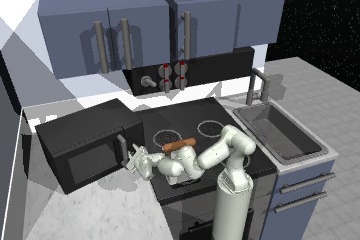}
        \includegraphics[width=0.130\textwidth]{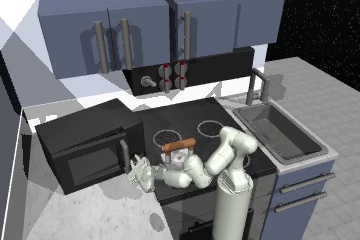}
        \includegraphics[width=0.130\textwidth]{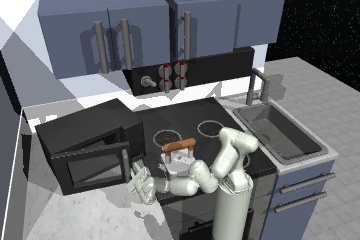}
        \includegraphics[width=0.130\textwidth]{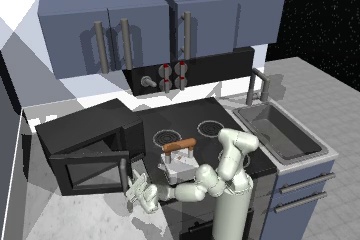}
        \\
        \includegraphics[width=\textwidth]{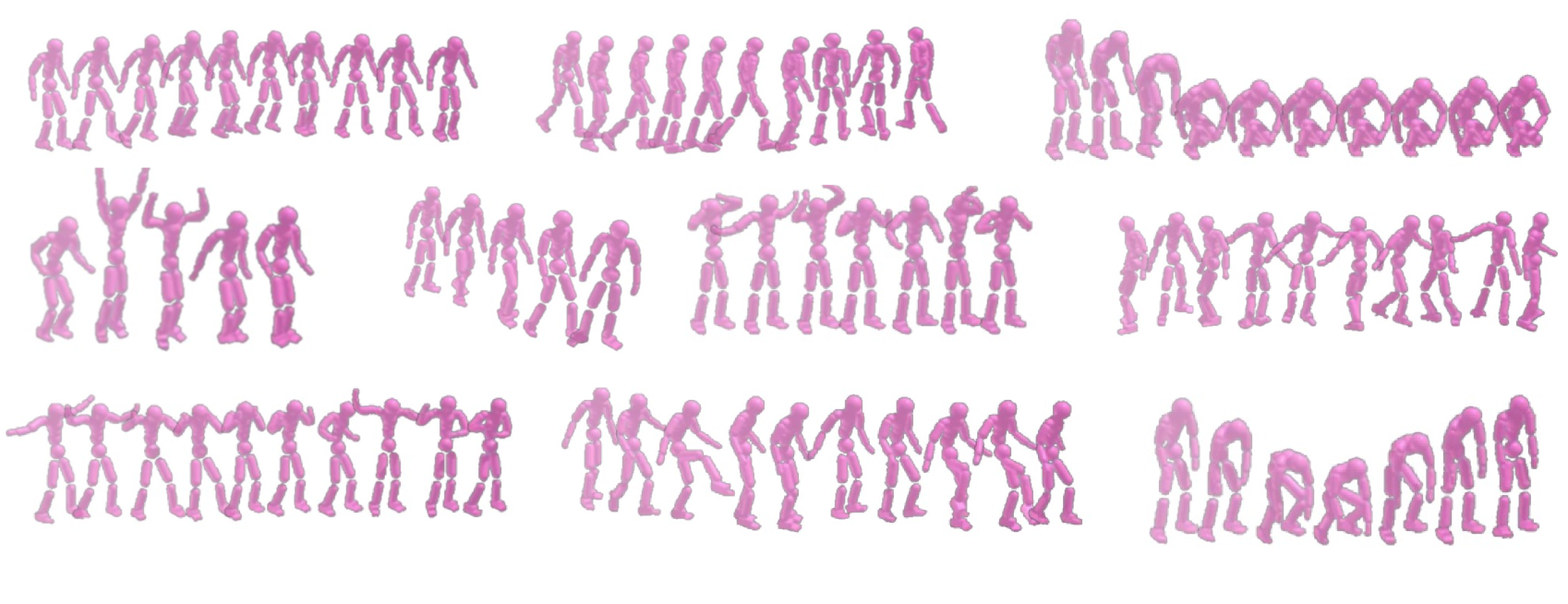}
        \vspace{-20pt}
	\caption{\textbf{More discovered diverse skills in three domains.} We can see that in the Push domain, blocks are pushed to different positions. In the Kitchen domain, the robotic arm executes distinct actions. In the Humanoid domain, the agent exhibits different movements and navigates in different directions (the color progression from light to dark indicates the movement progress of the humanoid).   
	} 
	\label{app:fig:diverse_skills}
\end{figure*}

In Figure~\ref{app:fig:diverse_skills}, we provide more discovered diverse skills in the Push, Kitchen, and Humanoid domains.  
By applying DIDI across different tasks, we observe its flexibility and robustness in learning. In the Push domain, the variation in block positions highlights DIDI's effectiveness in handling straightforward tasks. Moving to the Kitchen domain, the range of learned skills, from manipulation of objects to intricate interactions with the environment, further illustrates its versatility. Finally, in the Humanoid domain, the development of complex motor skills such as jumping and crouching underscores DIDI's potential in navigating and performing in high-dimensional, dynamic environments. 

\begin{figure*}[ht!]
        \begin{center}
        \includegraphics[width=0.32\textwidth]{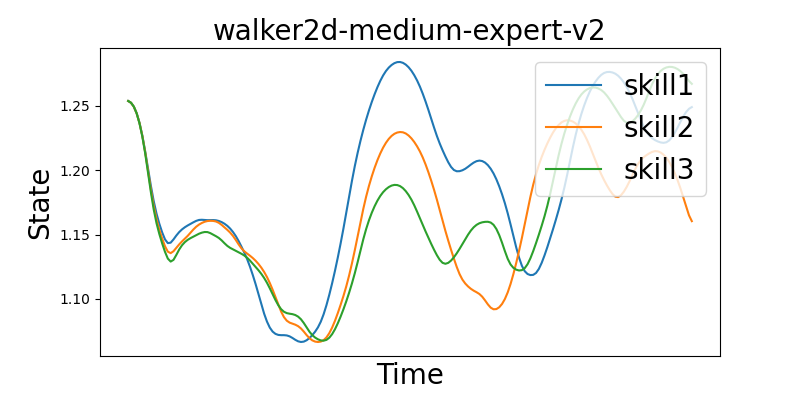}
        \includegraphics[width=0.32\textwidth]{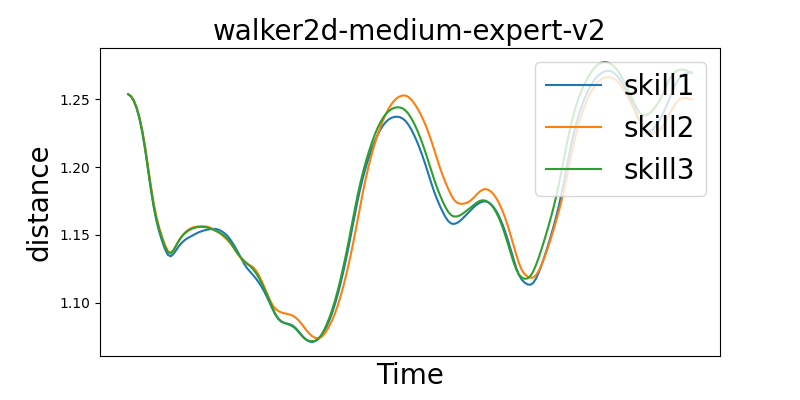}
        \includegraphics[width=0.32\textwidth]{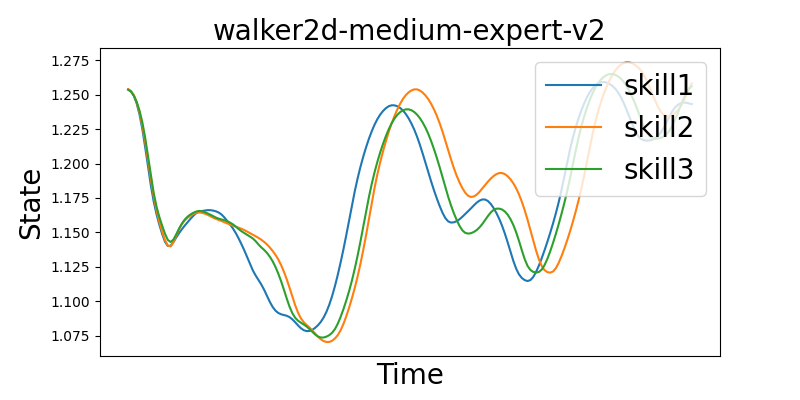}
        \includegraphics[width=0.32\textwidth]{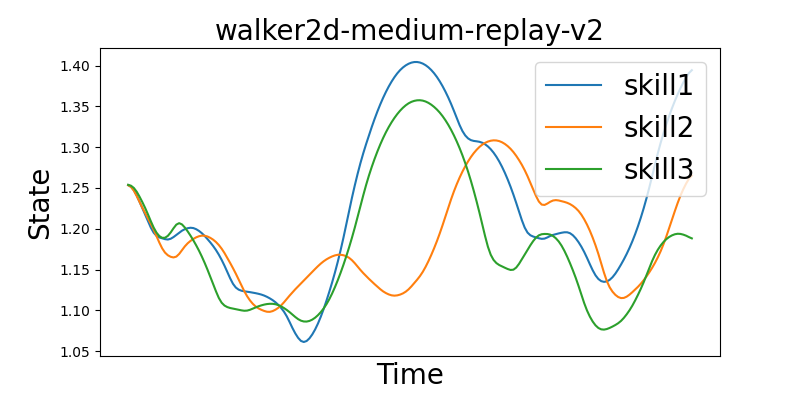}
        \includegraphics[width=0.32\textwidth]{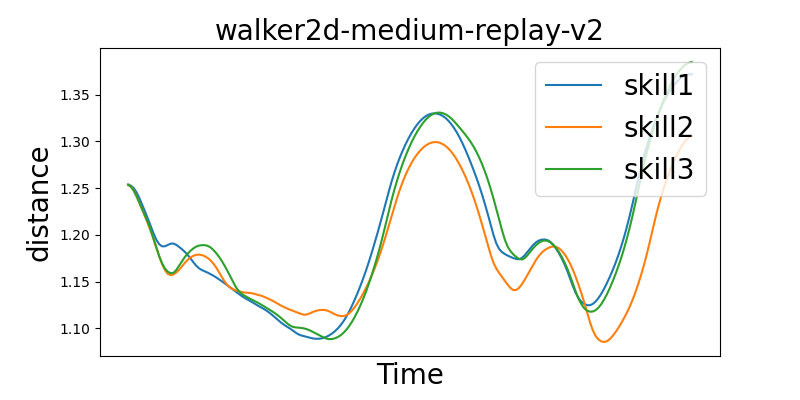}
        \includegraphics[width=0.32\textwidth]{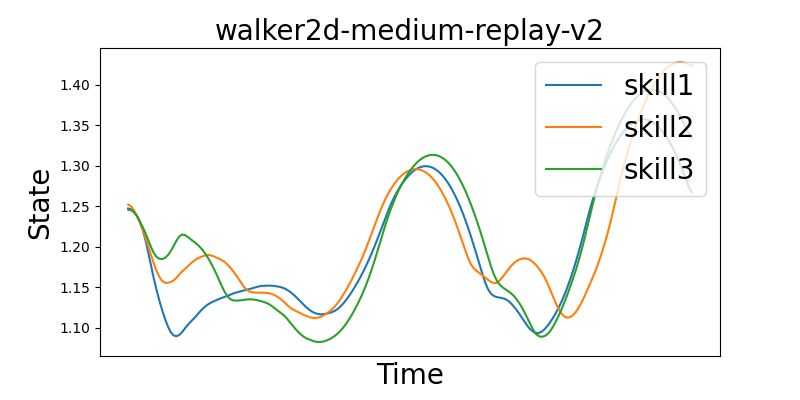}
        \includegraphics[width=0.32\textwidth]{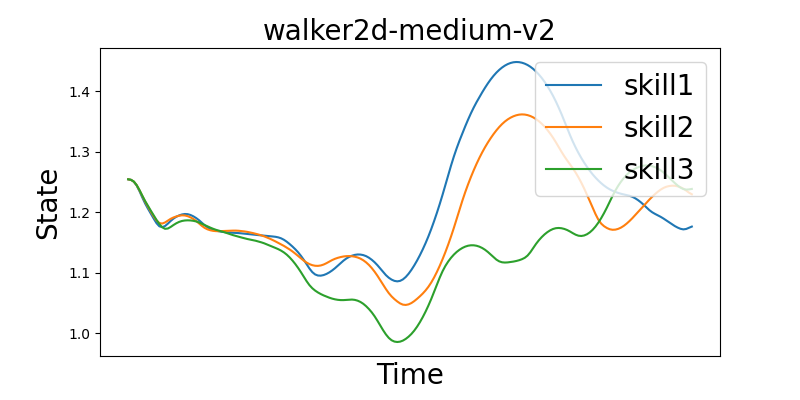}
        \includegraphics[width=0.32\textwidth]{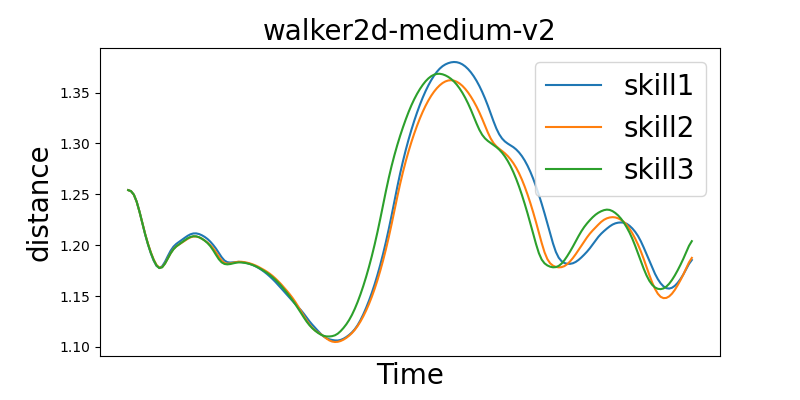}
        \includegraphics[width=0.32\textwidth]{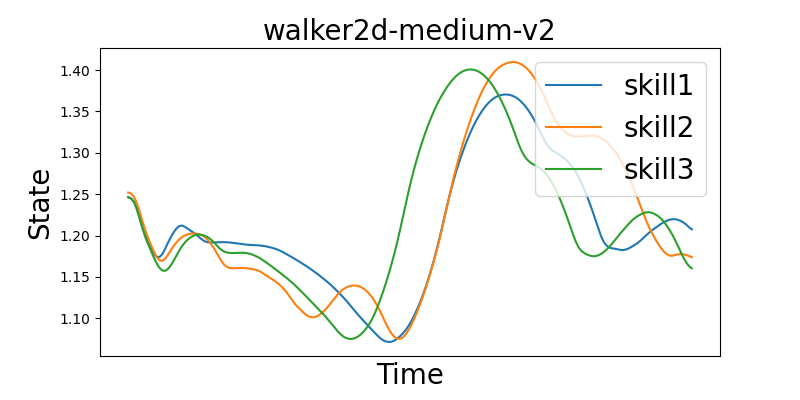} \\
        {DIDI} \qquad\qquad\qquad\qquad\qquad\qquad
        {VAE-DI} \qquad\qquad\qquad\qquad\qquad\qquad
        {Diffuser}
        \end{center}
        \vspace{-2pt}
        \caption{
        \textbf{Skill (Diversity) Visualization.}
        The figures illustrate the learned skills by DIDI, VAE-DI, and Diffuser on the D4RL walker2d domain. The skills demonstrated by DIDI exhibit a high level of diversity compared to VAE-DI and Diffuser. Each row represents a different task (walker2d-medium-expert-v2, walker2d-medium-replay-v2, walker2d-medium-v2), and each column compares the performance of the three methods. The Y-axis represents the state, and the X-axis represents time.
        }
	\label{appendix:fig:walker2d} 
        \vspace{-2pt}
\end{figure*}

\begin{figure*}[ht!]
        \begin{center}
        \includegraphics[width=0.32\textwidth]{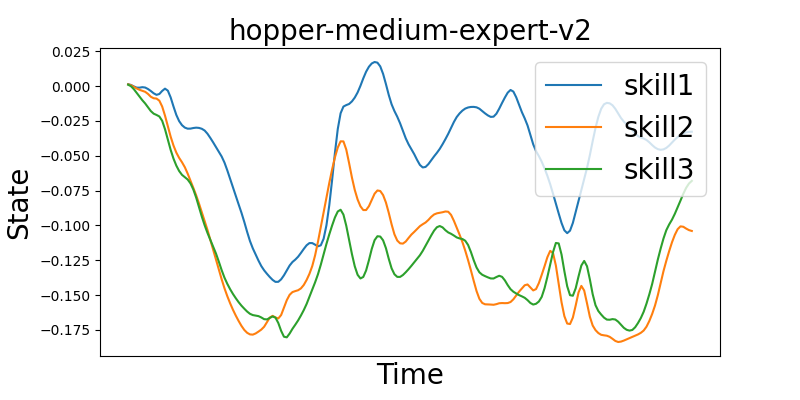}
        \includegraphics[width=0.32\textwidth]{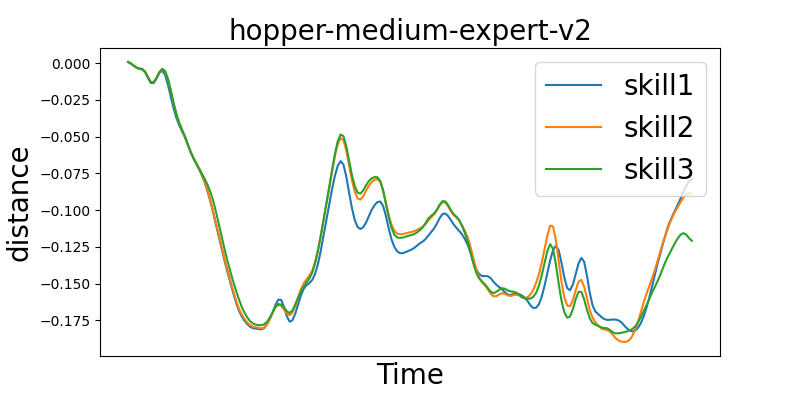}
        \includegraphics[width=0.32\textwidth]{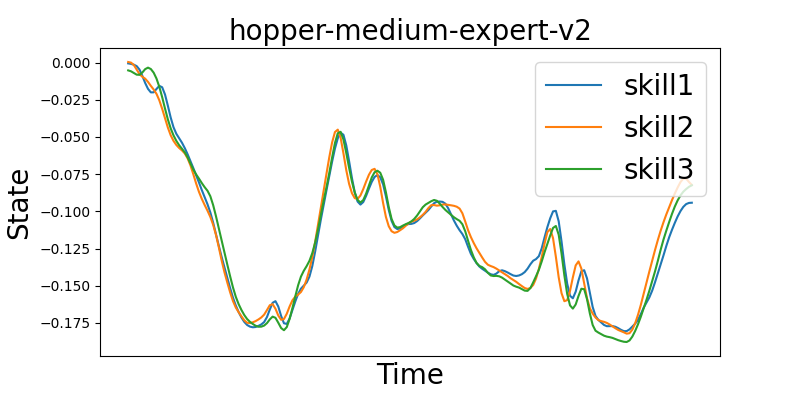}
        \includegraphics[width=0.32\textwidth]{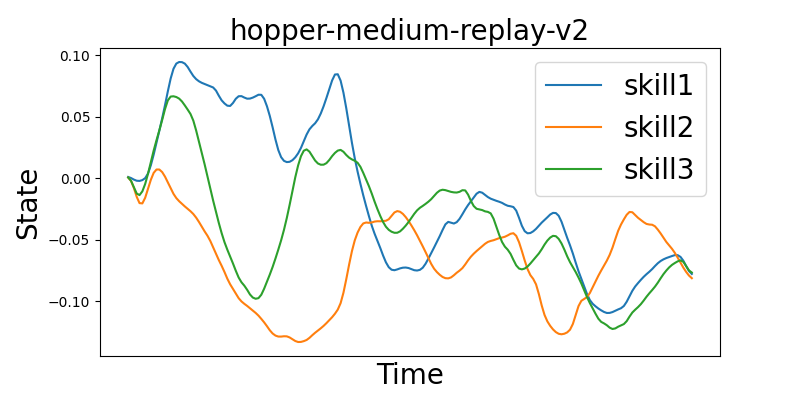}
        \includegraphics[width=0.32\textwidth]{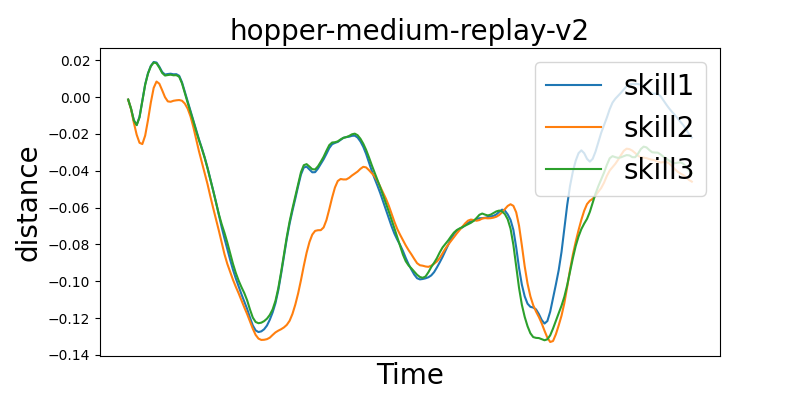}
        \includegraphics[width=0.32\textwidth]{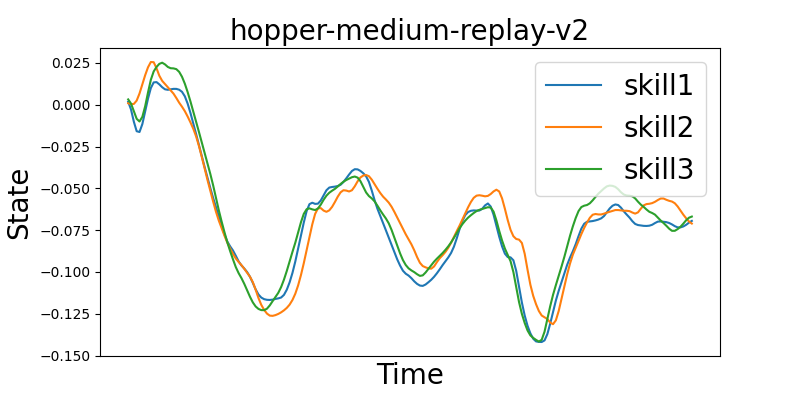}
        \includegraphics[width=0.32\textwidth]{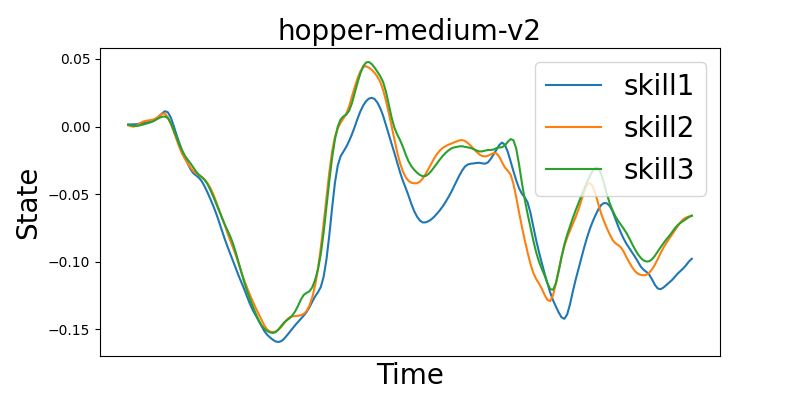}
        \includegraphics[width=0.32\textwidth]{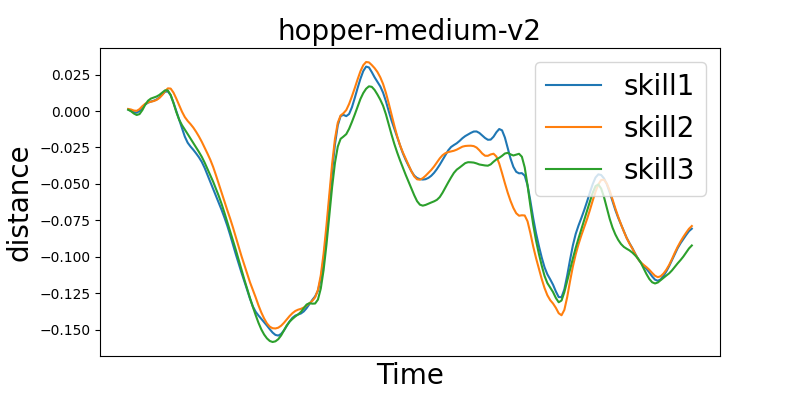}
        \includegraphics[width=0.32\textwidth]{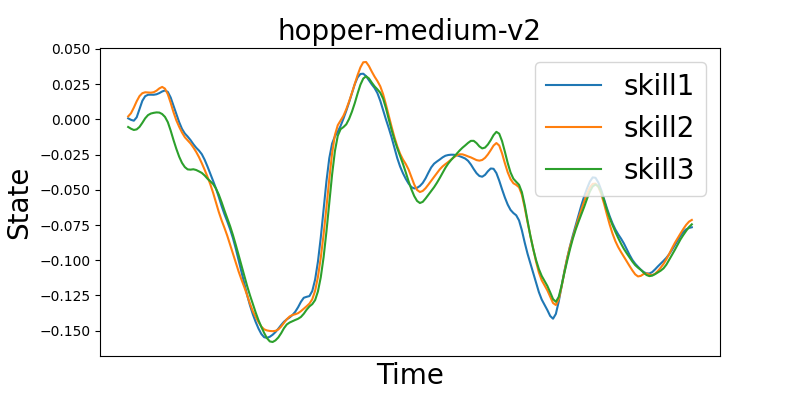}\\
        {DIDI} \qquad\qquad\qquad\qquad\qquad\qquad
        {VAE-DI} \qquad\qquad\qquad\qquad\qquad\qquad
        {Diffuser}
        \end{center}
        \caption{\textbf{Skill (Diversity) Visualization.}
        The figures illustrate the learned skills by DIDI, VAE-DI, and Diffuser on the D4RL hopper domain. The skills demonstrated by DIDI exhibit a high level of diversity compared to VAE-DI and Diffuser. Each row represents a different task (hopper-medium-expert-v2, hopper-medium-replay-v2, hopper-medium-v2), and each column compares the performance of the three methods. The Y-axis represents the state, and the X-axis represents time.
        }
	\label{appendix:fig:hopper} 
        \vspace{-2pt}
\end{figure*}

\begin{figure*}[ht!]
        \begin{center}
        \includegraphics[width=0.32\textwidth]{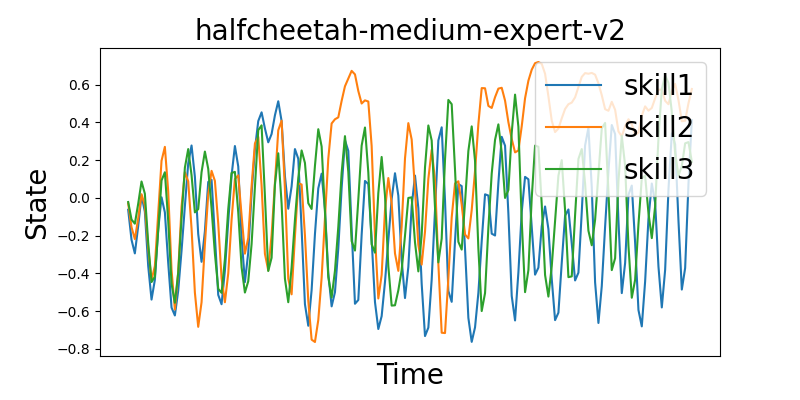}
        \includegraphics[width=0.32\textwidth]{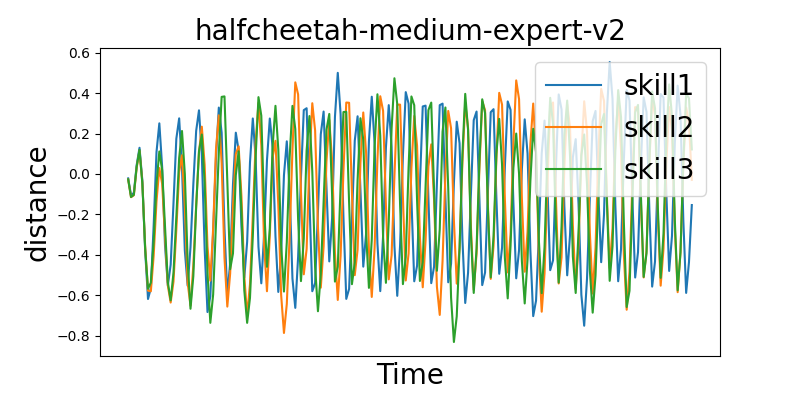}
        \includegraphics[width=0.32\textwidth]{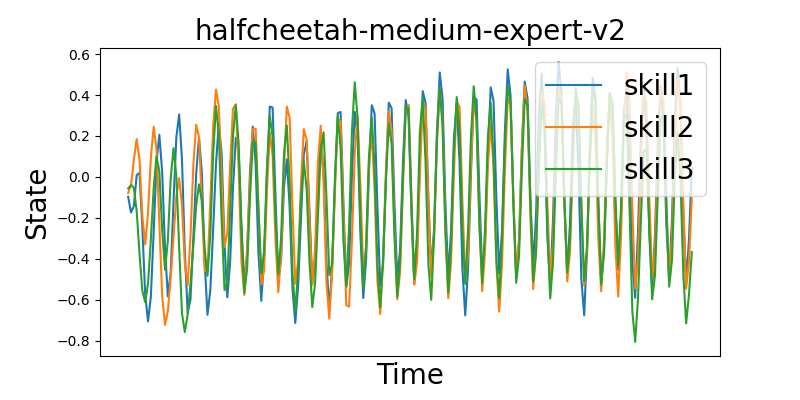}
        \includegraphics[width=0.32\textwidth]{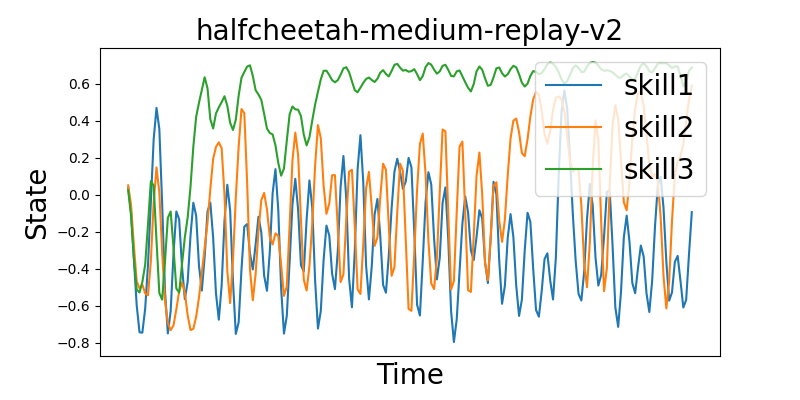}
        \includegraphics[width=0.32\textwidth]{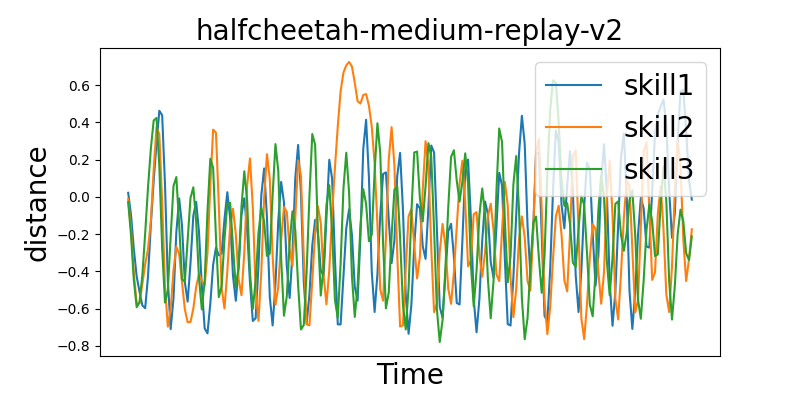}
        \includegraphics[width=0.32\textwidth]{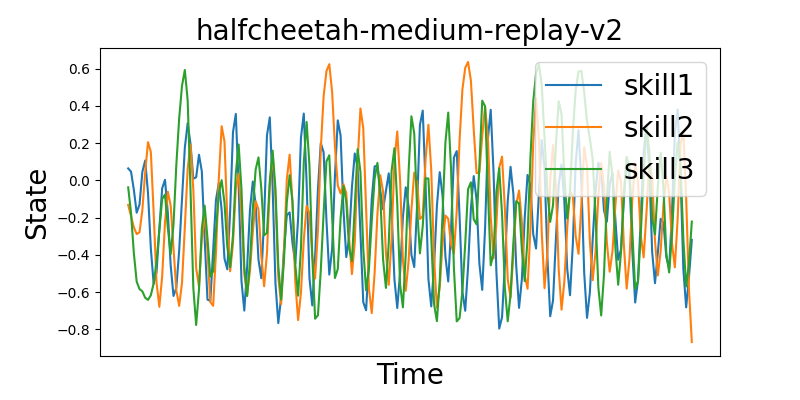}
        \includegraphics[width=0.32\textwidth]{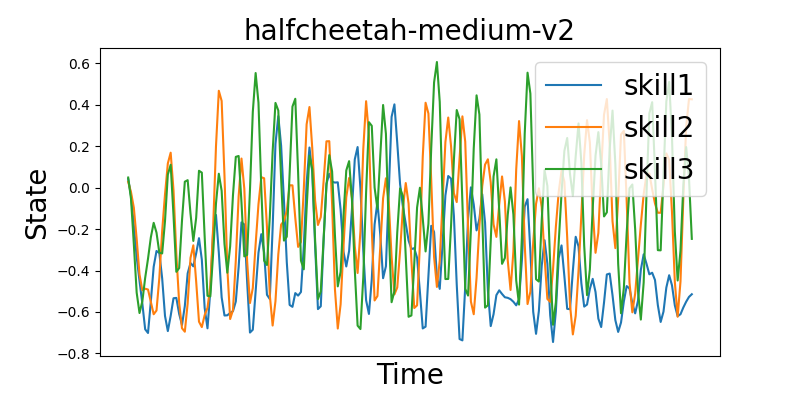}
        \includegraphics[width=0.32\textwidth]{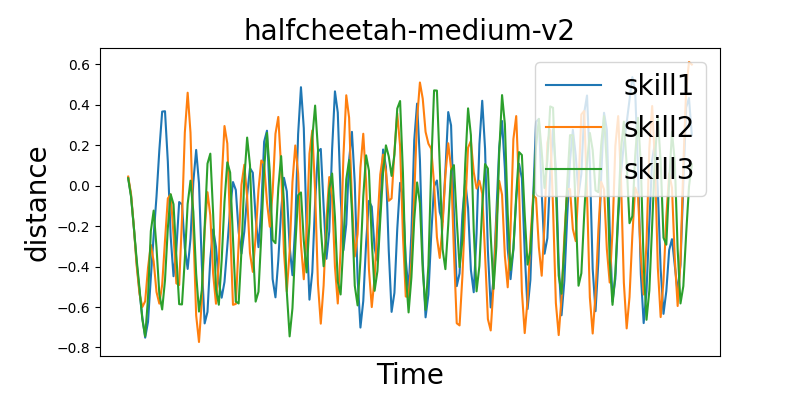}
        \includegraphics[width=0.32\textwidth]{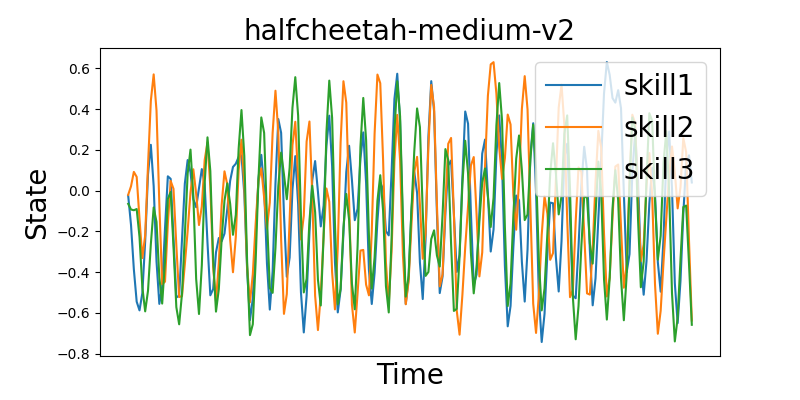}\\
        {DIDI} \qquad\qquad\qquad\qquad\qquad\qquad
        {VAE-DI} \qquad\qquad\qquad\qquad\qquad\qquad
        {Diffuser}
        \end{center}
        \caption{\textbf{Skill (Diversity) Visualization.}
        The figures illustrate the learned skills by DIDI, VAE-DI, and Diffuser on the D4RL halfcheetah domain. The skills demonstrated by DIDI exhibit a high level of diversity compared to VAE-DI and Diffuser. Each row represents a different task (halfcheetah-medium-expert-v2, halfcheetah-medium-replay-v2, halfcheetah-medium-v2), and each column compares the performance of the three methods. The Y-axis represents the state, and the X-axis represents time.
        }
	\label{appendix:fig:halfcheetah} 
        \vspace{-2pt}
\end{figure*}

In Figures~\ref{appendix:fig:walker2d},~\ref{appendix:fig:hopper} and~\ref{appendix:fig:halfcheetah}, we visualize the learned skills by DIDI, VAE-DI, and Diffuser on D4RL tasks. We can see that the learned skills (by DIDI) exhibit a high level of diversity (compared to VAE-DI and Diffuser). 
These results provide further quantitative comparisons, demonstrating that our approach achieves high levels of diversity while maintaining competitive performance with the guidance of extrinsic rewards. 


\newpage


\end{document}